\documentclass[article,shortnames,nojss]{jss}

\usepackage{thumbpdf,lmodern}

\usepackage{framed}
\usepackage{dsfont}
\usepackage{amsmath,amsthm}
\usepackage{amsfonts}
\usepackage{amssymb}
\usepackage[figuresright]{rotating}
\usepackage{orcidlink}
\usepackage{float} %
\usepackage{cleveref}
\usepackage{subcaption}

\theoremstyle{definition}

\newcommand{\python}{\proglang{Python}}
\newcommand{\rlang}{\proglang{R}}
\newcommand{\cpp}{\proglang{C\texttt{++}}}
\newcommand{\rust}{\proglang{Rust}}

\newcommand{\julia}{\proglang{Julia}}

\newcommand{\pytorch}{\pkg{PyTorch}}

\newcommand{\keras}{\pkg{Keras}}
\newcommand{\jax}{\pkg{JAX}}

\newcommand{\tidymodels}{\pkg{tidymodels}}
\newcommand{\torch}{\pkg{torch}}
\newcommand{\sklearn}{\pkg{scikit-learn}}
\newcommand{\flux}{\pkg{Flux}}
\newcommand{\luz}{\pkg{luz}}
\newcommand{\libtorch}{\pkg{LibTorch}}
\newcommand{\tensorflow}{\pkg{TensorFlow}}

\newcommand{\mlj}{\pkg{MLJ}}
\newcommand{\mlrt}{\pkg{mlr3}}
\newcommand{\mlrtpipelines}{\pkg{mlr3pipelines}}
\newcommand{\mlrttorch}{\pkg{mlr3torch}}

\DeclareMathAlphabet{\mathmybb}{U}{bbold}{m}{n}

\ifdefined\N                                                                %
\else
  \newcommand{\N}{\mathds{N}}
\fi
\ifdefined\C
  \renewcommand{\C}{\mathds{C}}                                             %
\else
  \newcommand{\C}{\mathds{C}}
\fi
\newcommand{\xv}{x}													%

\renewcommand{\P}{\mathds{P}}                                               %
\renewcommand{\xi}[1][i]{\xv^{(#1)}}                                          %

\author{Sebastian Fischer~\orcidlink{0000-0002-9609-3197} \\
    LMU Munich \\
    MCML \\
    \And Lukas Burk~\orcidlink{0000-0001-7528-3795} \\
    LMU Munich \\ MCML \\
    Leibniz Institute for\\Prevention Research and\\Epidemiology -- BIPS \\
    University of Bremen
    \AND Carson Zhang~\orcidlink{0009-0005-7730-4832} \\
    LMU Munich \\
    \And Bernd Bischl~\orcidlink{0000-0001-6002-6980} \\
    LMU Munich \\
    MCML \\
    \And Martin Binder~\orcidlink{0009-0008-2578-2869} \\
    LMU Munich \\
    MCML \\
}

\Plainauthor{Sebastian Fischer, Lukas Burk, Carson Zhang, Bernd Bischl, Martin Binder}

\title{\mlrttorch{}: A Deep Learning Framework in \rlang{} Based on \mlrt{} and \torch{}}
\Plaintitle{mlr3torch: A Deep Learning Framework in R Based on mlr3 and torch}
\Shorttitle{\mlrttorch{}: Deep Learning in \rlang{} Based on \mlrt{} and \torch{}}

\Abstract{
Deep learning (DL) has become a cornerstone of modern machine learning (ML) practice.
We introduce the \rlang{} package \pkg{mlr3torch}, a modular and extensible DL framework for the \pkg{mlr3} ecosystem that is built on the \pkg{torch} package and does not depend on \python{}.
Neural networks become regular \pkg{mlr3} learners that can be resampled, tuned, benchmarked, and combined with other ML methods in pipelines.
The package provides ready-to-use learners for classification and regression, including recent tabular architectures and pretrained image models for fine-tuning, and supports custom learning problems.
Images and other tensors are stored lazily as columns of a task, so that multimodal and larger-than-memory datasets can be preprocessed and augmented like tabular features.
Networks can be written as \pkg{torch} modules or defined in the graph language of \pkg{mlr3pipelines}, which represents preprocessing, augmentation, and the architecture in a single graph that can be tuned jointly.
We explain the package's modular design, in which all components can be combined and extended, and present three use cases: a simple neural architecture search, the fine-tuning of a pretrained image network, and a multimodal network.
Runtime benchmarks show that \pkg{mlr3torch} is close to a pure \pkg{torch} implementation, except for very small networks.
}

\Plainkeywords{deep learning, neural networks, machine learning, mlr3, torch, R}

\Address{
  Sebastian Fischer \\
  Institut f\"ur Statistik \\ Ludwig-Maximilians-Universit\"at M\"unchen \\
  Ludwigstr. 33, 80539 Munich, Germany \\
  E-mail: \email{sebastian.fischer@stat.uni-muenchen.de}
}

\usepackage{array}
\renewenvironment{CodeChunk}{\par\addvspace{0.6\baselineskip}\noindent\begin{minipage}{\linewidth}}{\end{minipage}\par\vspace{0.6\baselineskip}}

\begin{document}

\section{Introduction}

The recent impact of deep learning (DL) across various fields, including computer vision, natural language processing, and, more recently, tabular data \citep{ref-hollmann2025tabpfn, ref-erickson2025tabarena}, can hardly be overstated.
To train neural networks efficiently, one needs GPU-accelerated tensor libraries, as well as automatic differentiation.
However, many other mathematical algorithms also rely on tensor operations and gradients. Consequently, these toolkits are useful beyond the training of neural networks. One example of this is the fitting of Gaussian processes in the \pkg{GPyTorch} package \citep{gardner2018gpytorch}.

While most of the DL ecosystem is centered around \python{}, \rlang{} \citep{ref-R-base} also offers DL packages, both wrappers of \python{} libraries and packages built on the native \torch{} package \citep{ref-torch2025}.

\mlrt{} \citep{ref-mlr32019} is an object-oriented machine learning (ML) framework for \rlang{} that consists of a collection of packages and provides a unified interface to many learning algorithms, together with resampling, benchmarking, hyperparameter tuning, and ML pipelines \citep{ref-mlr3pipelines2021}.
The core idea of \mlrttorch{}\footnote{\url{https://github.com/mlr-org/mlr3torch}} is to extend this ecosystem with DL capabilities in a way that feels familiar to its user base.
Neural networks become regular \mlrt{} learners that can seamlessly replace other ML models and directly benefit from these features: they can be tuned via \pkg{mlr3tuning} \citep{ref-mlr3tuning2024}, evaluated with nested resampling, and benchmarked against other learners, e.g., on high-performance computing (HPC) clusters.

To this end, \mlrttorch{} offers ready-to-use learners for classification and regression on tabular data and images, including pretrained image networks for fine-tuning, as well as generic interfaces for custom learning problems.
Through the \code{lazy_tensor} type, tensors of arbitrary modalities can be stored and preprocessed like regular columns of a \code{Task}, even when they do not fit into memory.
This allows to naturally handle arbitrary modalities and multimodal data and bridges the gap between \mlrt{}, a tabular ML framework, and \torch{}, a tensor-based DL library.
Networks can be predefined, written as \torch{} modules, or expressed in the graph language of \mlrtpipelines{}, which represents preprocessing, augmentation, and the architecture in a single graph that can be tuned over.
Through its modular and customizable design, third-party packages can extend the package, e.g. to add support for new architectures or task types.
\Cref{tab:comparison} compares the package with other \rlang{} frameworks for training neural networks, \Cref{tab:overview} lists further details of its functionality, and \Cref{sec:distinguishing} discusses its distinguishing features.

We see four main user groups for \mlrttorch{}.
First, many statisticians and domain scientists already work with \rlang{} and \mlrt{} and want to apply DL without switching to another language.
This demand is also reflected in \rlang{} packages that implement DL for specific disciplines.
At the time of writing, more than 50 CRAN packages depend on \torch{}, covering, e.g., the analysis of satellite image time series in Earth observation \citep{ref-simoes2021sits}, forest dynamics in ecology \citep{ref-finn}, risk stratification from electronic health records \citep{ref-dicerclust}, stochastic mortality modeling \citep{ref-nnmomo}, and the forecasting of extreme events \citep{ref-eqrn}.

Second, DL is relevant for tabular data, which is the main focus of \mlrt{}.
While gradient-boosted trees remain strong on tabular data \citep{ref-borisov2022deep}, neural networks perform better on some datasets \citep{ref-mcelfresh2024neural}.
Moreover, ensembles that combine different model families advance the state of the art in tabular ML \citep{ref-erickson2025tabarena}.
In \mlrttorch{}, neural networks can be combined with, e.g., gradient-boosted trees via stacking in \mlrtpipelines{}.
The AutoML system \pkg{mlr3automl} \citep{ref-mlr3automl} already uses the tabular learners of \mlrttorch{} as candidates next to gradient-boosted trees and other learners, and selects and tunes them jointly.

Third, \mlrttorch{} offers infrastructure for developers of packages that provide DL methods for specific tasks or disciplines.
For example, \pkg{survdnn} \citep{ref-elbadisy2026survdnn}, discussed in \Cref{sec:dl-in-r}, implements its own cross-validation and tuning code for deep survival models.
If such a package is instead built on top of \mlrttorch{}, which its object-oriented design allows, many components of \mlrttorch{}, including most of the architectures, can be reused.
Package developers can therefore focus on the task-specific parts, such as the losses.

Fourth, DL is also relevant indirectly, for methodological research in which either a DL-specific or a model-agnostic method is being developed and evaluated.
Examples include statistical tests or confidence intervals for the generalization error or interpretability methods.
Access to DL models is thus required to enable such research in \rlang{}.

Overall, we expect \mlrttorch{} to be used primarily for research and data analysis by users who are familiar with \rlang{}, rather than for deploying models in production.
Nevertheless, trained networks can be exported via \proglang{TorchScript} (\Cref{sec:dl-in-r}).

In the following sections, we provide an overview of the package. We start by reviewing related software and comparing \mlrttorch{} with other \rlang{} packages for DL in~\Cref{sec:comparison}, where we also discuss how it relates to the \pytorch{} DL ecosystem.
Then, we give a brief introduction to DL and the \pkg{torch} package in~\Cref{sec:background}.
Next, \Cref{sec:building-blocks} summarizes the package's main building blocks. This includes the data representation, the \code{Learner} class, as well as the neural network \code{Graph} representation.
\Cref{sec:extending} shows how to extend the package. Then, we demonstrate the package's capabilities using three use cases in~\Cref{sec:use-cases}.
There, we cover hyperparameter tuning, fine-tuning, and the construction of neural networks for multimodal problems.
Afterwards, we compare the runtime of \mlrttorch{} with that of \torch{} and \pytorch{} in~\Cref{sec:benchmarks}.
Finally, we summarize the package and discuss future development in~\Cref{sec:conclusion}.

\section[mlr3torch in the DL ecosystem]{\mlrttorch{} in the DL ecosystem}\label{sec:comparison}

\subsection{DL libraries and ML frameworks}\label{sec:dl-libraries}

The default language for DL is \python{} \citep{ref-van1995python} with its libraries \tensorflow{} \citep{ref-abadi2016tensorflow}, \pytorch{} \citep{ref-pytorch}, and \pkg{JAX} \citep{ref-jax2018github}.
Initially, \tensorflow{} exclusively supported static execution, which relies on building up a computational graph that is then optimized and executed. This is great for deployment because it allows for more optimizations before execution. In contrast, \pytorch{} became popular because of its dynamic approach, which is often called eager execution. This allows for a simpler mental model because less is happening behind the scenes, comes with more flexibility, and allows for easier debugging.
Nowadays, \tensorflow{} also supports eager execution, while \pytorch{} also allows for static compilation, which shows that both approaches have their strengths.
Nonetheless, \pytorch{} is arguably the de facto standard for DL as of today.
More recently, \pkg{JAX} introduced what can be described as a functional code transformation framework for \pkg{NumPy} programs \citep{ref-harris2020array}, with the gradient being one such transformation.
Another noteworthy \python{} library is \pkg{tinygrad} \citep{ref-tinygrad}, which strives to be a lightweight framework that can easily be extended to new hardware platforms.
These libraries focus primarily on the core building blocks of DL and mostly leave other functionality, such as training loops, preprocessing, or data augmentation, to other frameworks.
One example of such a framework is \pkg{Keras} \citep{ref-chollet2018keras}, which offers a more high-level application programming interface (API) that is both integrated into \tensorflow{} and available as its own multi-backend library with support for \pytorch{}, \tensorflow{}, and \pkg{JAX}.
The graph language of \mlrttorch{} was originally inspired by the graph-building interface from \pkg{Keras}.
Furthermore, \pkg{PyTorch Lightning} \citep{ref-lightning2019} and \pkg{fastai} \citep{howard2020fastai} offer higher-level frameworks for \pytorch{}, and \pkg{Flax} \citep{ref-flax} offers similar capabilities for \pkg{JAX}.
Languages other than \python{} also have DL libraries.
These include \julia{} \citep{ref-bezanson2017julia} with \flux{} \citep{ref-innes2018flux}, \rust{} \citep{ref-matsakis2014rust} with its crates \pkg{Burn} \citep{ref-burn} and \pkg{Candle} \citep{ref-candle}, and \proglang{Go} \citep{ref-go} with \pkg{GoMLX} \citep{ref-gomlx}.
In \cpp{} there is \pkg{LibTorch}, the \cpp{} backend for \pytorch{}, which can also be used directly in \cpp{}.

In addition to specialized DL toolkits, there are other, more general-purpose ML libraries that primarily focus on tabular data.
In \python, there is \sklearn{} \citep{ref-pedregosa2011scikit-learn}, in \julia{} there is \mlj{} \citep{ref-blaom2020mlj}, and \rlang{} has \tidymodels{} \citep{ref-kuhn2020tidymodels} and \pkg{mlr3} \citep{ref-mlr32019} with their respective predecessors \pkg{caret} \citep{ref-kuhn2008building} and \pkg{mlr} \citep{ref-bischl2016mlr}, both of which are still widely used.
From the perspective of these general ML frameworks, neural networks are just one type of learning algorithm, next to other methods like decision trees or support vector machines.
Connector packages exist that integrate DL methods into their generic APIs.
For example, the \pkg{skorch} package \citep{ref-skorch} integrates \pytorch{} into \sklearn{}, and \pkg{MLJFlux} \citep{ref-MLJFlux} does the same for \flux{} and \mlj{}; the corresponding \rlang{} packages for \tidymodels{} are discussed in the next section.

\subsection[DL in R]{DL in \rlang{}}\label{sec:dl-in-r}

Currently, there are two main routes for DL in \rlang{}: either using a \python{} library via \pkg{reticulate} \citep{ref-reticulate2025}, or using \torch{} \citep{ref-torch2025} or one of the packages built on top of it.
First, we discuss packages wrapping \python{} libraries and then cover native \rlang{} packages.

\paragraph{Wrappers of \python{} libraries}
\pkg{keras3} \citep{ref-keras32025} and \pkg{fastai} \citep{ref-fastai}, as well as the lower-level \pkg{tensorflow} \citep{ref-r-tensorflow2024} package, wrap their \python{} counterparts via \pkg{reticulate} \citep{ref-reticulate2025}.
They therefore inherit the large feature set of these libraries, which goes beyond what the \rlang{}-native alternative \torch{} currently offers.
For example, \pkg{keras3} can use \tensorflow{} \citep{ref-abadi2016tensorflow}, \jax{} \citep{ref-jax2018github}, or \pytorch{} \citep{ref-pytorch} as its backend and offers preprocessing layers for text and audio data as well as weight quantization.
Moreover, models can be trained on multiple GPUs or TPUs, e.g., via the distribution API of \keras{}, which is currently only available for the \jax{} backend.
For model serving, \pkg{keras3} models can be exported to formats such as SavedModel or ONNX, which can then be deployed.
In contrast, for exporting complete models for use outside of \rlang{}, \torch{} only supports \proglang{TorchScript}, which can be loaded from other languages such as \python{} or \cpp{}, but has been deprecated by \pytorch{} in favor of \code{torch.export}\footnote{\url{https://docs.pytorch.org/docs/2.9/jit.html}}.
Note that the latest release of the \rlang{} wrapper of \pkg{fastai} dates from April 2024.

However, bridging two interpreted languages also has some costs.
First, a \python{} environment with library versions that are compatible with the \rlang{} wrapper is required.
Although recent versions of \pkg{reticulate} can create such environments automatically, users have to manage a second set of dependencies next to their \rlang{} packages.
Moreover, \pkg{reticulate} can only bind to a single \python{} environment per \rlang{} session.
As a consequence, two \rlang{} packages that wrap \python{} libraries with conflicting version requirements cannot be used in the same session.
This runs counter to the philosophy of CRAN, where the latest \rlang{} packages should work together.
Similarly, because \rlang{} and \python{} libraries are loaded into the same process, their compiled dependencies can conflict.
Second, data has to be converted between the two languages, which comes with additional conversion overhead.
Third, the two languages follow different conventions: \python{} uses 0-based indexing, while \rlang{} uses 1-based indexing.
\rlang{} packages that wrap \python{} libraries, such as \pkg{keras3}, convert indices for many of their functions, but not for all.
Fourth, memory is managed by two separate garbage collectors.
Because \rlang{}'s garbage collector does not know how much memory a referenced \python{} object occupies, large objects can be kept alive longer than necessary, which can, e.g., exhaust the GPU memory unless the garbage collector is called manually.
While \torch{} tensors are also not visible to \rlang{}'s garbage collector, \torch{} mitigates this by triggering it when the memory usage exceeds a threshold or an allocation fails \citep{ref-torch2025}.
Finally, errors that occur in the \python{} code come with \python{} tracebacks, and \rlang{}'s debugging tools cannot step into the \python{} code.

\paragraph{\rlang{}-native tensor libraries}
\torch{}, which is built directly on top of \pkg{LibTorch} and works similarly to \pytorch{}, provides tensors with GPU support, automatic differentiation capabilities, neural network modules, and optimizers.
It is the foundation of various DL packages discussed below.
There is also the minimal \rlang{} package \pkg{tinytorch}, which offers bindings to \libtorch{} \citep{tinytorch}, but it currently only supports Linux and is not available on CRAN.
A different approach is taken by \pkg{anvl} \citep{ref-anvl}, which is inspired by \jax{}.
It compiles \rlang{} functions into optimized programs via the XLA compiler \citep{ref-openxla}, has support for automatic differentiation, and can also run programs on NVIDIA GPUs.
However, \pkg{anvl} focuses on accelerated tensor computing in general rather than on DL, and does not provide neural network layers or optimizers.
As such, it provides the computational base layer for DL, but none of the higher-level infrastructure.

\paragraph{\rlang{}-native training frameworks and model packages}
\luz{} \citep{ref-luz2025}, \pkg{cito} \citep{ref-cito2024}, and \pkg{kindling} \citep{ref-soetewey2026kindling} are higher-level training frameworks for \torch{}.
Because these are closest in spirit to \pkg{mlr3torch}, \Cref{tab:comparison} contains an in-depth comparison of the four packages (\luz{}~0.5.2, the development version 1.2 of \pkg{cito} on GitHub (commit \code{aec71d5}), \pkg{kindling}~0.3.2, and \mlrttorch{}~0.4.0), all of which run on the CPU and on GPUs and require no \python{}.
\luz{} encapsulates the training loop into reusable pieces, such as callbacks and metrics, and makes few assumptions about the training setup, which makes it the best fit when maximum flexibility is needed.
The callback mechanism of \mlrttorch{} was partly inspired by the implementation in \luz{}.
\pkg{cito} is a focused package that allows fitting multi-layer perceptrons (MLPs) through a formula interface and convolutional neural networks on arrays or image folders, and can combine both via \code{mmn()}. It is easy to use when it covers one's specific use case, but has limited flexibility.
\pkg{kindling} is designed to integrate \pkg{torch} into \pkg{tidymodels} and thus benefits from the latter's ecosystem.
While it supports arbitrary data modalities, its \pkg{tidymodels} integration is restricted to tabular modeling.
Both \pkg{cito} and \pkg{kindling} prioritize ease of use, as training is configured through the arguments of a single function call.
\mlrttorch{} takes a different approach and decomposes the model and its training into modular components.

\begin{sidewaystable}
\centering
\footnotesize
\setlength{\tabcolsep}{4pt}
\renewcommand{\arraystretch}{1.0}
\newcommand{\note}[1]{{\scriptsize(#1)}}
\begin{tabular}{@{}>{\raggedright\arraybackslash}p{0.16\textheight}*{4}{>{\raggedright\arraybackslash}p{0.195\textheight}}@{}}
\hline
 & \luz{} & \pkg{cito} & \pkg{kindling} & \mlrttorch{} \\
\hline
Framework integration & -- & -- & \tidymodels{} \note{tabular only} & \mlrt{} \\
Task types & arbitrary & classif., (multi-output) regr., count, multi-label, custom \note{via custom loss} & classif., (multi-output) regr., custom \note{via custom loss} & classif., regr., custom \\
Data modalities & arbitrary \note{\code{torch::dataset}} & tabular, fixed-size arrays or image folders \note{e.g., images; up to 3D} & arbitrary \note{single input tensor; \tidymodels{} integration only for tabular data} & arbitrary \\
Preprocessing & in the \code{dataset} & model formula \note{tabular only} & \pkg{recipes} \citep{ref-recipes} \note{tabular only}; in the \code{dataset} & Via (many predefined or custom) \pkg{PipeOp}s \\
Data augmentation & in the \code{dataset}; mixup callback & 3 built-in or custom & in the \code{dataset} & Via (many predefined or custom) \pkg{PipeOp}s \\
Multimodal inputs & \checkmark  & (\checkmark) \note{several tabular and array inputs via \code{mmn()}} & -- & \checkmark \\
Tabular architectures & -- & MLP & MLP \note{plus recurrent networks} & 5 \note{see \Cref{tab:overview}} \\
Pretrained image models & -- & 20 \note{\pkg{torchvision}} & -- & 46 \note{\pkg{torchvision}} \\
Custom architectures & \code{nn_module} & \code{nn_module} \note{\code{custom} argument}; CNN layer lists \note{\code{create\_architecture()}} & \code{nn_module} \note{as layer of \code{nn\_arch()}} & \code{nn_module}, graph language \\
Early stopping metric & any metric & loss & loss & any metric \\
Custom callbacks & \checkmark{} \note{25 stages, incl. prediction} & (\checkmark) \note{after every epoch; \code{dnn()} only} & -- & \checkmark{} \note{12 stages, training only; predefined ones in \Cref{tab:overview}} \\
Resuming training & \checkmark{} \note{incl. optimizer and callback states} & (\checkmark) \note{optimizer and scheduler re-initialized} & -- & \checkmark{} \note{incl. optimizer and callback states} \\
Third-party extensions & \checkmark{} \note{callbacks, metrics; built-ins can be subclassed} & (\checkmark) \note{custom losses, modules, augmentations as arguments} & (\checkmark) \note{custom layers, activations, losses as arguments} & \checkmark{} \note{all component types, used like built-in ones} \\
Performance measures & 8; custom via \code{luz_metric()} & -- \note{loss only} & many via \pkg{yardstick} \citep{ref-yardstick}, custom & 50+ \mlrt{} \code{Measure}s; custom \\
Hyperparameter tuning & -- & random search \note{tabular only} & \pkg{tune} \citep{ref-tune} \note{tabular only} & \pkg{mlr3tuning} \\
Tunability & -- & selected architecture and training settings & selected architecture and training settings & all hyperparameters, incl.\ network topology \\
Resampling & -- \note{train-validation split only} & k-fold CV for tuning \note{tabular only} & via \pkg{rsample} \citep{ref-rsample} \note{tabular only; nested via \pkg{nestedtune} \citep{ref-nestedtune}} & many schemes, incl.\ nested resampling \\
Parallelization & -- & tuning \note{CPU cores, GPUs}, bootstrapping \note{CPU cores}; tabular only & via \tidymodels{} \note{\pkg{future} \citep{ref-future}, \pkg{mirai} \citep{ref-mirai}; tabular only} & \pkg{future}, HPC via \pkg{mlr3batchmark} \\
Interpretability & -- & built-in, model-agnostic and gradient-based & built-in, model-specific; model-agnostic via \tidymodels{} & model-agnostic via \mlrt{} packages \note{e.g., \pkg{xplainfi} \citep{ref-burk2026xplainfi}} \\
\hline
\end{tabular}
\caption{Comparison of general-purpose \rlang{} frameworks for training neural networks with \torch{}.
\checkmark: supported; (\checkmark): possible with limitations; --: not directly available.}
\label{tab:comparison}
\end{sidewaystable}

\paragraph{Model packages}
Furthermore, there are various packages that offer predefined models and building blocks for \torch{}.
The main focus here is on tabular neural networks.
The \pkg{tabnet} \citep{ref-r-tabnet} package implements the attention-based TabNet architecture for tabular data \citep{ref-arik2021tabnet}.
\pkg{brulee} \citep{ref-brulee2025} implements various tabular architectures, such as SAINT \citep{ref-somepalli2021saint} or the pretrained tabular foundation model TabICLv2 \citep{ref-qu2026tabiclv2}.
It is also integrated into the \tidymodels{} ecosystem.
It offers a set of easy-to-use functions with a fixed set of options exposed as function arguments.
This is convenient when the package covers the use case at hand but has limited flexibility and customization options.
The recently released \pkg{tabby} package \citep{ref-tabby} integrates tabular neural networks from different \rlang{} packages into \tidymodels{} by registering them as \pkg{parsnip} \citep{ref-parsnip} model engines, currently those of \pkg{brulee} and \pkg{tabpfn} \citep{ref-tabpfn-r, ref-hollmann2025tabpfn}, where the latter wraps the \python{} implementation of TabPFN via \pkg{reticulate}.
It thus serves as a unifying interface for tabular neural networks within the \tidymodels{} framework.
Within the \mlrt{} ecosystem, \pkg{mlr3extralearners} \citep{ref-mlr3extralearners} provides learners for the tabular foundation models TabPFN \citep{ref-hollmann2025tabpfn} and TabFM \citep{ref-tabfm}; like \pkg{tabpfn}, these learners wrap the respective \python{} implementations via \pkg{reticulate}.
Finally, it is also possible to use \pkg{keras3} models with \tidymodels{} via the \pkg{kerasnip} \citep{ref-kerasnip} integration package.

Several other packages exist beyond these \tidymodels{} integrations.
For survival analysis, \pkg{survdnn} \citep{ref-elbadisy2026survdnn}, which is built directly on \torch{}, offers MLPs with different survival losses for tabular data with a formula interface, including cross-validation and hyperparameter tuning via grid search.
Pretrained models are provided, e.g., by \pkg{torchvision} \citep{ref-r-torchvision}, which implements pretrained image models, image datasets, and image transformations, and by \pkg{minhub} \citep{ref-minhub}, which provides minimal \torch{} implementations of pretrained language models such as Llama or Gemma, whose weights can be loaded from the Hugging Face Hub.
Finally, \pkg{h2o} \citep{ref-h2o} trains feed-forward networks and autoencoders without GPU support, and the older packages \pkg{nnet}, \pkg{neuralnet}, \pkg{deepnet}, and \pkg{RSNNS} \citep{ref-nnet, ref-neuralnet, ref-deepnet, ref-rsnns} implement small networks on the CPU.

\subsection[Distinguishing features of mlr3torch]{Distinguishing features of \mlrttorch{}}\label{sec:distinguishing}

Because \mlrttorch{} is built on \mlrt{}, neural networks are regular learners that profit from the whole ecosystem, including resampling, a large collection of performance measures, hyperparameter tuning, and model-agnostic interpretability methods.
Compared with the \tidymodels{} integrations discussed above, \mlrt{} in particular offers first-class nested resampling (a tuned learner is itself a learner and can be passed directly to resampling or benchmarking functions)\footnote{In \tidymodels{}, nested resampling is available via the separate \pkg{nestedtune} package \citep{ref-nestedtune}, which is under active development and not yet available on CRAN.}, large-scale benchmarks that can easily be run on high-performance computing clusters via \pkg{mlr3batchmark} \citep{ref-mlr3batchmark} and \pkg{batchtools} \citep{ref-batchtools2017}, and a wider range of tuning methods \citep{ref-mlr3tuning2024, ref-mlr3mbo, ref-mlr3hyperband}.
In addition, the number of epochs can be tuned internally via early stopping, and this internal tuning can be combined with the offline tuning of the remaining hyperparameters (\Cref{sec:learner-torch,sec:tuning}).
This mechanism was implemented in \mlrt{} as a general mechanism for iterative learners, with \mlrttorch{} as one of its main motivating use cases, and is also used, e.g., by gradient-boosted trees \citep{ref-mlr3book-valid}.

Beyond this integration, \mlrttorch{} combines several features that are not offered together by the other \rlang{} packages.
First, images and other tensors are stored as \code{lazy_tensor} columns of a \code{Task}, next to standard tabular features and without being loaded into memory.
This makes multimodal datasets and datasets that do not fit into memory first-class citizens, and should feel familiar to \rlang{} users who work with data frames or \code{data.table}s.
In contrast, the \torch{}-based integrations into \tidymodels{} are restricted to tabular data, and non-tabular data can only be passed to \pkg{keras3} models via \pkg{kerasnip}, as arrays that are held in memory.
Tensor columns are preprocessed and augmented by the same kind of \code{PipeOp}s as tabular features; these transformations are applied lazily when a batch is loaded, and augmentation only during training.
Second, preprocessing, the network architecture, the training configuration, and ensembling such as stacking are all expressed in a single graph language.
As every node exposes its hyperparameters, not only layer widths or learning rates but also the topology of the network, e.g., the choice between alternative branches or the number of repetitions of a block, can be tuned jointly with the preprocessing (\Cref{sec:tuning}), and components can simply be added, removed, or exchanged for ablation studies.
The graph also defines how the input tensors of each network entry point are constructed from the \code{Task}, which makes multi-input architectures straightforward (\Cref{sec:multimodal}).
Third, users can choose their level of control, from ready-to-use learners over arbitrary \torch{} modules converted to learners to custom architectures in the graph language.
Fourth, all components, from network layers and preprocessing operators to losses, callbacks, and task types, are represented by classes with well-defined interfaces from which users can inherit, rather than by a fixed set of options exposed as function arguments.
Other packages can therefore register additional components and reuse the infrastructure of both \mlrt{} and \mlrttorch{} instead of reimplementing it.
For example, \pkg{mlr3torchAUM} \citep{ref-mlr3torchAUM} adds losses for AUC optimization and batch samplers, \pkg{mlr3tuningspaces} \citep{ref-mlr3tuningspaces} provides published search spaces for the tabular learners of \mlrttorch{}, and \pkg{mlr3automl} \citep{ref-mlr3automl} includes these learners in its AutoML system.
Finally, \mlrttorch{} requires only native \rlang{} packages and does not depend on any \python{} libraries.
Taken together, these features make \mlrttorch{} a research toolbox for DL in \rlang{}.

\begin{table}[p]
\centering
\small
\begin{tabular}{@{}>{\raggedright\arraybackslash}p{0.2\textwidth}>{\raggedright\arraybackslash}p{0.76\textwidth}@{}}
\hline
Aspect & Functionality in \mlrttorch{} \\
\hline
Task types & Classification and regression out of the box; one-off custom learning problems available via the generic \code{TaskTorch}. Third-party packages can add support for new task types (\Cref{sec:extending-learner-task}). \\
Data modalities & Tabular features and generic tensors via the \code{lazy_tensor} type (\Cref{sec:data-representation}). Text data is not supported out of the box; tokenization needs to be performed by a third-party package such as \pkg{tok} \citep{ref-tok}. Multimodal architectures combine several entry points of a network graph, e.g., for images and tabular features (\Cref{sec:multimodal}). \\
Preprocessing and augmentation & 11 preprocessing and 14 augmentation operators, most of them based on \pkg{torchvision} \citep{ref-r-torchvision}. Tabular data can be preprocessed via functionality from \pkg{mlr3pipelines} \citep{ref-mlr3pipelines2021}. Custom preprocessing and augmentation methods can be easily added via \code{pipeop_preproc_torch()}. \\
Predefined architectures & Tabular: MLP, residual network (TabResNet) and FT-Transformer \citep[both][]{ref-gorishniy2021revisiting}, TabM \citep{ref-gorishniy2025tabm}, and RealMLP \citep{ref-holzmuller2024realmlp}. Image: 46 architectures from \pkg{torchvision}, including pretrained weights. \\
Custom architectures & Arbitrary \code{nn_module}s via \code{LearnerTorchModule} (\Cref{sec:extending-learner-task}) or the graph language (\Cref{sec:graphs}), which offers more than 70 layer operators, including convolutional, pooling, normalization, attention, transformer, and recurrent layers, an operator for arbitrary tensor functions, and repeated blocks; custom layers can be created via \code{pipeop_torch()}. \\
Training & Losses (cross-entropy, MSE, L1), optimizers (SGD, Adam, AdamW, RMSprop, Adagrad) using the fast ``ignite'' implementations of \torch{} written in \cpp{} (\Cref{sec:benchmarks}), CPU or GPU training, and the number of epochs as an internally tuned hyperparameter. Predefined callbacks support logging (e.g., to TensorBoard), checkpointing (which allows resuming the training), learning rate scheduling, and fine-tuning. Custom callbacks, losses, and optimizers can be created via \code{torch_callback()}, \code{as_torch_loss()}, and \code{as_torch_optimizer()}; callbacks can hook into 12 stages of the training loop. \\
Evaluation and tuning & All \mlrt{} performance measures, which can also be tracked during training and used for early stopping; custom ones via \code{Measure} classes or \code{msr_torch()}. Resampling, including nested resampling, and benchmarking. Tuning methods such as random search, Bayesian optimization, or Hyperband for all hyperparameters, including the network topology and, internally via early stopping, the number of epochs. Model-agnostic interpretability via \mlrt{} packages such as \pkg{xplainfi} \citep{ref-burk2026xplainfi}. Stacking with other learners and parallel or HPC execution via \pkg{future} \citep{ref-future} and \pkg{mlr3batchmark} \citep{ref-mlr3batchmark}. \\
Extensibility & Learners, pipeline operators, callbacks, losses, and optimizers are registered in dictionaries and task types via \mlrt{}'s reflections, so components from third-party packages are used exactly like built-in ones (\Cref{sec:extending}). \\
\hline
\end{tabular}
\caption{Details of the functionality of \mlrttorch{}, complementing the comparison with other packages in \Cref{tab:comparison}.}
\label{tab:overview}
\end{table}

\subsection[Relation to the PyTorch ecosystem]{Relation to the \pytorch{} ecosystem}\label{sec:relation-pytorch}

Because \torch{} and \pytorch{} are both built on \libtorch{}, most of the core functionality of \pytorch{} is also available in \rlang{}.
Models can be trained on CPUs, NVIDIA GPUs, and Apple Silicon GPUs.
However, some functionality of \pytorch{} is not available in \torch{}, and we list the key differences below.
First, \torch{} does not provide bindings for the distributed API of \libtorch{}, so data-parallel training across multiple GPUs or nodes is not supported.
Second, the just-in-time (JIT) compiler \code{torch.compile()} \citep{ref-ansel2024pytorch} is implemented in \python{} and operates on \python{} bytecode, which is why it cannot be used from \rlang{}.
Nonetheless, \torch{} supports \proglang{TorchScript}, the legacy JIT compiler of \pytorch{}.
Third, custom GPU kernels are harder to write in \torch{}, because the \python{}-based kernel language \proglang{Triton} \citep{ref-tillet2019triton} is not available.
Instead, one has to resort to \proglang{CUDA} and \cpp{}.
Fourth, the composable function transforms of \pytorch{}, such as \code{vmap()}, \code{jacrev()}, or \code{hessian()}, are not available in \rlang{}.
Fifth, \torch{} can be slower than \pytorch{}, especially for small networks, where the overhead per operation outweighs the actual computation.
The missing support for distributed training and compilation mainly matters when training very large models.
We therefore do not expect large-scale DL to happen in \rlang{}, and \mlrttorch{} does not aim to compete in this space.

Otherwise, the main difference is that model implementations from the \python{} ecosystem cannot be used directly from \rlang{}.
However, because the API of \torch{} closely mirrors that of \pytorch{}, architectures that are not yet available in \rlang{} can usually be ported with relatively little effort by translating their \pytorch{} code.
Pretrained weights from the Hugging Face Hub can also be reused, as the safetensors format in which they are commonly stored can be read via the \pkg{safetensors} package \citep{ref-safetensors}.

\section{Background}\label{sec:background}

DL is a subfield of ML that uses artificial neural networks to model relationships in data.
These are highly expressive, overparameterized models that do not rely on manual feature engineering but instead learn the feature representation in an end-to-end manner.
The simplest architecture is the MLP, which combines affine-linear transformations with nonlinear activation functions:

\begin{equation}\label{eq:mlp}
f_{\theta}(\mathbf{x})
= \phi_{L}\!\bigl(
  W_{L}\,\phi_{L-1}\!\bigl(
    W_{L-1}\,\dotsm \phi_{1}(W_{1}\mathbf{x}+b_{1}) \dotsm + b_{L-1}
  \bigr) + b_{L}
\bigr)
\end{equation}

\noindent with $W_{\ell}\in\mathbb{R}^{d_{\ell}\times d_{\ell-1}}$, $b_{\ell}\in\mathbb{R}^{d_{\ell}}$, $\theta = \{W_1, \ldots, W_L, b_1, \ldots, b_L\}$ and $\phi_{\ell}: \mathbb{R}^{d_{\ell}} \rightarrow \mathbb{R}^{d_{\ell}}$ for $\ell=1,\dots,L$.

\begin{figure}[t!]
\centering
\includegraphics[width=\textwidth]{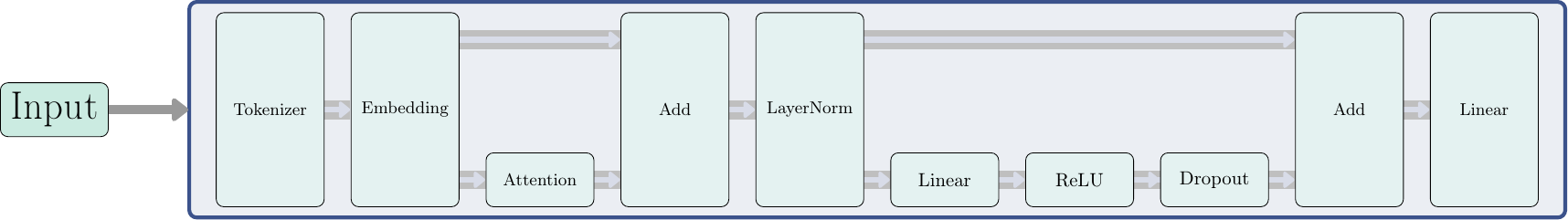}
\caption{Simple transformer-based neural network architecture represented as a graph.}
\label{fig:intro-network}
\end{figure}

Almost all neural network architectures can be represented as directed acyclic graphs (DAGs).
Each node in the graph represents a -- possibly parameterized -- transformation that takes in one or more tensors as input and outputs one or more tensors.
Edges go from layer outputs to layer inputs, defining the data flow between operations.
One such graph, a simple transformer-based neural network, is depicted in \Cref{fig:intro-network}.
First, the input (e.g., a text string) is split into tokens. These tokens are then converted to vector representations through an embedding layer.
The output tensor is then fed into an attention module \citep{vaswani2017attention} and subsequently added to the output of the embedding layer, in a technique commonly called a ``skip connection'' \citep{he2016deep}.
Note that the addition layer has no learnable parameters.
The output is then processed using layer normalization \citep{ba2016layernormalization}, followed by a linear layer, a rectified linear unit (ReLU) activation \citep{nair2010rectified, glorot2011deep}, and a dropout layer \citep{srivastava2014dropout}, the output of which is again added to the output of the normalization layer, creating another skip connection.
The final linear layer is the network head.

With $f_\theta$ denoting the whole network, where $\theta$ represents the union of all the parameters of the individual layers, the objective of training a neural network is to find parameters $\theta^*$ from the parameter space $\Theta$ that minimize a loss function $\mathcal{L}$ given a data-generating process $\P$.

\begin{equation}
\theta^* \in \operatorname{argmin}_{\theta \in \Theta} \mathbb{E}_{(\mathbf{x}, y) \sim \P} \left[ \mathcal{L}(f_{\theta}(\mathbf{x}), y) \right]
\end{equation}

In practice, this solution is approximated by minimizing the empirical loss on a dataset $\mathcal{D} = \{ (\mathbf{x}^{(i)}, y^{(i)}) \}_{i=1}^n$ sampled from $\P$.
Typically, optimization is performed using variants of stochastic gradient descent (SGD), which update the parameters in the direction of the negative gradient of the loss, estimated using a mini-batch of the data.
In iteration $t$, SGD draws a mini-batch $B_t \subset \{1, \ldots, n\}$, computes the gradient
\begin{equation}
g_t = \nabla_\theta \frac{1}{|B_t|} \sum_{i \in B_t} \mathcal{L}\bigl(f_{\theta_t}(\mathbf{x}^{(i)}), y^{(i)}\bigr), \label{eq:gradient}
\end{equation}
and updates the parameters as $\theta_{t+1} = \theta_t - \eta \, g_t$ with learning rate $\eta > 0$.
Basic SGD is often extended with weight decay, which shrinks the parameters towards zero in every update, acting as a regularizer.
Furthermore, many optimizers use momentum, where the parameter update is based on an exponentially decaying sum of past gradients.
One popular method is Adam (adaptive moment estimation) \citep{ref-kingma2015adam}, which additionally adjusts the step size of each weight using estimates of the first and second moments of its gradients.
In this paper, we mostly use AdamW \citep{ref-loshchilov2017decoupled}, a variant of Adam that decouples the weight decay from the adaptive update, which often improves generalization.

For these computations, \pkg{mlr3torch} builds on the \rlang{} package \pkg{torch} \citep{ref-torch2025}, which provides the building blocks for each of the steps described above.
Its central data structure is the \code{torch_tensor}, a multi-dimensional array that resides on a device, such as the CPU or a GPU, and which represents the features $\mathbf{x}^{(i)}$, the targets $y^{(i)}$, and the parameters $\theta$.
Because training neural networks mainly consists of large matrix multiplications and other operations that can be executed in parallel, GPUs can speed up training by orders of magnitude compared to CPUs, which is crucial for larger models and datasets.
The network $f_\theta$ itself has class \code{nn_module}, which holds the learnable parameters and defines the forward computation.
It is created by an object of class \code{nn_module_generator}, which is defined via \code{nn_module()} and which returns a module with initialized parameters when it is called.
Modules are composable, which allows the creation of reusable components for commonly used layers.
When the loss is computed, the operations on tensors that require gradients are recorded, so that the gradient $g_t$ from \Cref{eq:gradient} can be computed automatically via backpropagation.
For the optimization routines, \pkg{torch} provides the \code{torch_optimizer} class with various implementations, such as \code{optim_sgd} or \code{optim_adamw}.
Finally, a \code{dataset} defines how to retrieve a single observation $(\mathbf{x}^{(i)}, y^{(i)})$ of $\mathcal{D}$ as tensors, while a \code{dataloader} iterates over a \code{dataset} and combines (shuffled) observations into the mini-batches $B_t$.
This abstraction is needed because DL datasets are often too large to fit into memory, so mini-batches are loaded one at a time and the whole dataset is never kept in memory at once.
In \pkg{torch}, these building blocks are combined in a training loop that users write themselves:
for each epoch, the \code{dataloader} yields the mini-batches, and for each mini-batch, the network computes its predictions, the loss is calculated, its gradient is computed via backpropagation, and the optimizer updates the parameters.
Further functionality, such as tracking the performance on validation data or early stopping, also has to be implemented as part of this loop.
For a comprehensive introduction to \pkg{torch}, we refer to \citet{ref-keydana2023}.

We do not cover the theoretical foundations of DL in depth, because the topic is very broad and most readers are already familiar with the basic concepts. For more thorough coverage, please refer to \citet{bishop2024deep}.

\section{Building blocks}\label{sec:building-blocks}

Besides \pkg{torch} (\Cref{sec:background}), \pkg{mlr3torch} builds upon the \rlang{} packages \pkg{mlr3} \citep{ref-mlr32019} and \pkg{mlr3pipelines} \citep{ref-mlr3pipelines2021}, which we now briefly introduce.
Note that these are not specific to DL, but because \pkg{mlr3torch}'s object hierarchy builds upon their ontology, we introduce them first.

\pkg{mlr3} is a modern, feature-rich, and extensible framework for various ML tasks in the \rlang{} programming language, including regression and classification, as well as survival analysis or clustering through the extensions \pkg{mlr3proba} \citep{sonabend2021mlr3proba} and \pkg{mlr3cluster} \citep{ref-mlr3cluster}.
The ecosystem relies on \pkg{R6} \citep{ref-r6chang} for most user-facing objects and hence uses reference semantics.
At its core, \pkg{mlr3} allows access to and application of many different ML algorithms implemented across different \rlang{} packages through a unified interface.
This interface is built around a few core \pkg{R6} classes.
A \code{Task} contains the available data in tabular form, with the observations as rows and the features and the target as columns, as well as some metadata describing the modeling problem, such as which column is the target.
In \Cref{sec:data-representation} we use a predefined image classification task, and in \Cref{sec:finetuning} we show how to create one.
A \code{Learner} encapsulates an ML algorithm, which includes a structured representation of its hyperparameters, capability annotations, as well as the central methods \code{\$train()} and \code{\$predict()}, where the former trains a model and the latter uses the stored model to create a \code{Prediction} object.
From the perspective of \mlrt{}, the neural networks of \mlrttorch{} are thus simply \code{Learner}s; see \Cref{sec:learner-torch}.
The quality of predictions is assessed by a \code{Measure}, such as the classification error used in \Cref{sec:learner-torch}.
To estimate the generalization performance of a \code{Learner}, a \code{Resampling} defines how the data is split, e.g., via cross-validation.
In \Cref{sec:learner-torch}, we show how to run such a resample experiment.
To compare several \code{Learner}s on one or more tasks, \code{benchmark()} can be used, as we demonstrate in \Cref{sec:multimodal}.
Hyperparameter tuning is provided by the \pkg{mlr3tuning} extension \citep{ref-mlr3tuning2024}, which we use in \Cref{sec:tuning}.
Solving real-world modeling problems usually requires not only individual learners but also the creation of complete modeling workflows consisting of multiple steps.
For this, the \mlrtpipelines{} extension can be used.
It allows representing a learning pipeline as a \code{Graph} that consists of different \code{PipeOp}s.
Such \code{PipeOp}s can, e.g., represent a preprocessing, model training, or postprocessing step.
Just like the \code{Learner} class, \code{PipeOp}s have a \code{\$train()} and a \code{\$predict()} method.
Each \code{PipeOp} has one or more inputs and outputs.
Preprocessing steps for tabular features, such as scaling or encoding, are \code{PipeOp}s that inherit from \code{PipeOpTaskPreproc}.
A \code{Graph} can be converted into a \code{GraphLearner}, which can then be used like any other \code{Learner}.
For a more thorough coverage of \pkg{mlr3} and its ecosystem, see the freely accessible \pkg{mlr3} book \citep{ref-mlr3book}.

\begin{figure}[t]
\centering
\includegraphics[width=\textwidth]{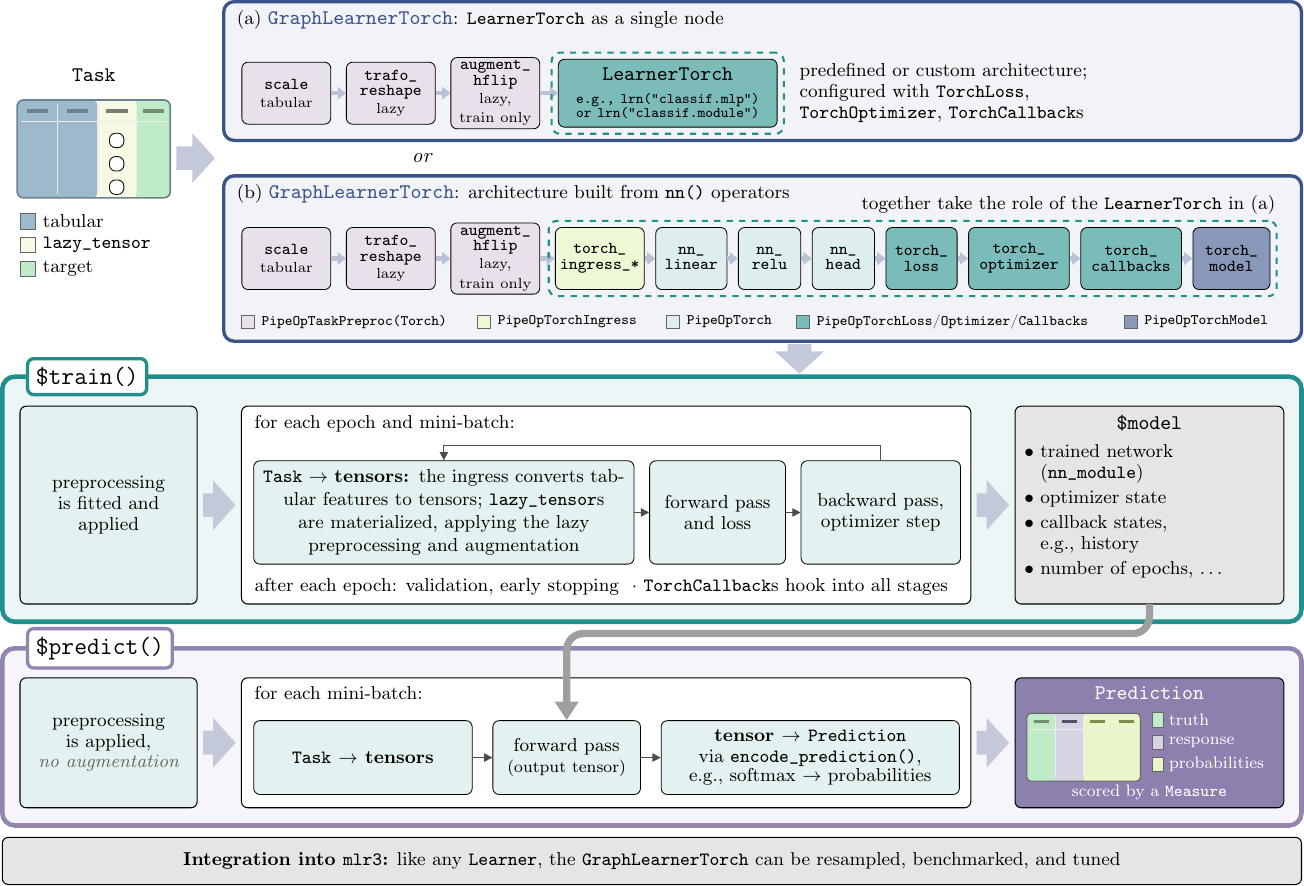}
\caption{Overview of the \mlrttorch{} workflow. A \code{GraphLearnerTorch} combines preprocessing and augmentation operators with either (a) a \code{LearnerTorch} as a single node or (b) an architecture built from \code{nn()} operators, where the operators from \code{torch\_ingress\_*} to \code{torch\_model} take the role of the \code{LearnerTorch}. In each mini-batch, a subset of the \code{Task}'s rows is converted to one or more tensors: the ingress converts tabular features, and \code{lazy\_tensor}s are materialized, which applies the lazy preprocessing and augmentation. Augmentation is only applied during training. The \code{\$model} contains the trained network and further training state, and during prediction, the network output is converted into a \code{Prediction} via \code{encode\_prediction()}.}
\label{fig:workflow}
\end{figure}

\begin{table}[t]
\centering
\small
\begin{tabular}{@{}ll>{\raggedright\arraybackslash}p{0.4\textwidth}@{}}
\hline
Category & Class & Role \\
\hline
Data & \code{mlr3::Task} & Data and metadata of an ML problem \\
 & \code{mlr3torch::TaskTorch} & \code{Task} for custom learning problems \\
 & \code{mlr3torch::lazy_tensor} & Column of a \code{Task} or \code{data.table} that holds tensors lazily \\
 & \code{torch::torch_tensor} & In-memory tensor on a device \\
 & \code{torch::dataset} & Access to individual observations \\
 & \code{torch::dataloader} & Draws mini-batches from a \code{dataset} \\
Preprocessing & \code{mlr3pipelines::PipeOp} & Single step of a \code{Graph} \\
 & \code{mlr3pipelines::Graph} & Pipeline of connected \code{PipeOp}s \\
 & \code{mlr3pipelines::PipeOpTaskPreproc} & Preprocesses a \code{Task}, e.g., its tabular features \\
 & \code{mlr3torch::PipeOpTaskPreprocTorch} & Preprocesses \code{lazy_tensor}s \\
Architecture & \code{torch::nn_module_generator} & Creates \code{nn_module}s \\
 & \code{torch::nn_module} & Network (or network segment) with its parameters \\
Training & \code{mlr3::Learner} & Learns and predicts from a \code{Task} \\
 & \code{mlr3pipelines::GraphLearner} & \code{Learner} that wraps a \code{Graph} \\
 & \code{mlr3torch::GraphLearnerTorch} & \code{GraphLearner} for a \code{Graph} that contains a neural network \\
 & \code{mlr3torch::LearnerTorch} & \code{Learner} that trains an \code{nn_module} \\
 & \code{torch::torch_optimizer} & Parameter update rule \\
 & \code{mlr3torch::TorchLoss} & Wraps a \pkg{torch} loss function \\
 & \code{mlr3torch::TorchOptimizer} & Wraps a \code{torch::torch_optimizer} \\
 & \code{mlr3torch::TorchCallback} & Customizes the training loop \\
Evaluation & \code{mlr3::Prediction} & Output of \code{Learner\$predict()} \\
 & \code{mlr3::Measure} & Scores a \code{Prediction} \\
 & \code{mlr3torch::MeasureTorch} & \code{Measure} for a \code{TaskTorch} \\
 & \code{mlr3::Resampling} & Data splits for \code{resample()} \\
\hline
\end{tabular}
\caption{Main classes of \pkg{mlr3}, \pkg{mlr3pipelines}, \pkg{torch}, and \mlrttorch{} for each step of the DL workflow. The classes for defining neural networks as \code{Graph}s are listed in \Cref{tab:graph-classes}.}
\label{tab:classes}
\end{table}

{\widowpenalty=10000 %
Building on these frameworks, \mlrttorch{} follows the usual \mlrt{} workflow.
Before we describe it in more detail in the remainder of this section, we first provide a high-level overview.
As in \pkg{mlr3}, the modeling problem is represented via a \code{Task}.
Because a \code{Task} usually stores its data in a \code{data.table}, in which every column is a vector, standard tabular features such as numbers or factors can be stored directly, but images or other tensors cannot.
For this reason, \mlrttorch{} introduces the \code{lazy_tensor} type, which is essentially a vector of tensors that can be stored as a column of a \code{data.table} (\Cref{sec:data-representation}).
It is lazy in the sense that it does not hold the tensors in memory but only describes how to obtain them, e.g., by loading images from disk.
Only when they are needed are the tensors created as \code{torch_tensor}s.
This abstraction allows \mlrttorch{} to go beyond tabular modeling problems and support arbitrary data modalities.
Similarly, \mlrttorch{} goes beyond the classification and regression tasks of \mlrt{}: custom learning problems can be represented via the generic \code{TaskTorch} class and evaluated with the generic \code{MeasureTorch} class, which inherit from \code{Task} and \code{Measure}, respectively (\Cref{sec:extending-learner-task}).\par}

A \code{lazy_tensor} can also be preprocessed lazily: transformations and data augmentation are implemented as \code{PipeOpTaskPreprocTorch} objects (a subclass of \code{PipeOpTaskPreproc}) that do not modify the tensors directly but only record the transformation, which is applied when the tensors are materialized (\Cref{sec:data-representation}).
Thus, \mlrttorch{} offers a unified interface for preprocessing standard tabular features and generic tensors.

Neural networks are trained by a \code{LearnerTorch}, whose \code{\$train()} method implements the training loop described in \Cref{sec:background} and whose \code{\$predict()} method applies the trained network to mini-batches of new data and then encodes the resulting tensors as a \code{Prediction} object.
A \code{LearnerTorch} is configured by a loss, an optimizer, and callbacks (\Cref{sec:learner-torch}).
These are represented by \code{TorchLoss}, \code{TorchOptimizer}, and \code{TorchCallback} objects, which wrap the corresponding \pkg{torch} objects and make their hyperparameters available to \mlrt{}, e.g., for tuning.
\mlrttorch{} offers different levels of control over the network architecture of a \code{LearnerTorch}.
The simplest options are predefined architectures, such as the MLP \code{LearnerTorchMLP}, for which only architectural hyperparameters, such as the number and width of the layers, are configurable (\Cref{sec:learner-torch}).
Custom architectures can be defined as a \code{nn_module_generator} and turned into a \code{Learner} via \code{LearnerTorchModule} (\Cref{sec:extending-learner-task}).
Finally, the architecture can be expressed in the graph language of \mlrtpipelines{}, where each layer is a \code{PipeOpTorch} (\Cref{sec:graphs}).
As a \code{LearnerTorch} is an \mlrt{} \code{Learner}, it can be resampled, benchmarked, and tuned like any other \code{Learner}.
\Cref{fig:workflow} gives an overview of the workflow.
Preprocessing and augmentation operators can be combined with a \code{LearnerTorch} in a \code{Graph}, which is converted via \code{as_learner_torch()} into a \code{GraphLearnerTorch}, the subclass of \code{GraphLearner} that \mlrttorch{} provides for graphs containing a neural network (\Cref{fig:workflow}(a)).
Alternatively, the architecture and the training configuration can be expressed as operators in the same \code{Graph} (\Cref{fig:workflow}(b) and \Cref{fig:pipeop-torch}).
In both cases, the hyperparameters of all components can be tuned jointly, and in the latter case, this includes the structure of the architecture (\Cref{sec:tuning}).
\Cref{tab:classes} gives an overview of the classes.

\subsection{Data representation}\label{sec:data-representation}

\subsubsection[The lazy tensor type]{The \code{lazy\_tensor} type}\label{sec:lazy-tensor}\nopagebreak

To support not only standard tabular features in a \code{Task} (\Cref{sec:building-blocks}), but also generic \torch{} tensors, \pkg{mlr3torch} defines the \code{lazy_tensor} type, which represents a vector of tensors with possibly heterogeneous shapes.
It is built upon the \code{torch::dataset} class, and therefore only needs to define how to obtain its elements, but not necessarily store them in memory -- hence the name ``lazy''.
We demonstrate this data type using the MNIST digit classification example problem \citep{ref-mnist-2012}, a supervised learning task where the goal is to categorize an image of a handwritten digit into one of the ten classes (0--9).
To make the results in this paper reproducible, we set the seeds of both \rlang{}'s and \pkg{torch}'s random number generators. For more information on reproducibility, see \Cref{app:comp-details}.

\begin{CodeChunk}
\begin{CodeInput}
R> library("mlr3torch")
R> set.seed(42)
R> torch_manual_seed(42)
R> mnist <- tsk("mnist")
R> mnist
\end{CodeInput}
\begin{CodeOutput}
-- <TaskClassif> (70000x2): MNIST Digit Classification -------------------
* Target: label
* Properties: multiclass
* Features (1):
  * lt (1): image
* Target classes: 1 (11%), 7 (10%), 3 (10%), 2 (10%), 9 (10%), 0 (10%), 6
(10%), 8 (10%), 4 (10%), 5 (9%)
\end{CodeOutput}
\end{CodeChunk}

The underlying data includes two columns, one of which is a \code{factor} vector containing the class labels, and the other is a \code{lazy_tensor} containing grayscale images with $28\times28$ pixels.
\filbreak

\begin{CodeChunk}
\begin{CodeInput}
R> rows <- mnist$data(1:2)
R> rows
\end{CodeInput}
\begin{CodeOutput}
    label           image
   <fctr>   <lazy_tensor>
1:      5 <tnsr[1x28x28]>
2:      0 <tnsr[1x28x28]>
\end{CodeOutput}
\end{CodeChunk}

We can call the \code{materialize()} function to convert a \code{lazy_tensor} to a list of \code{torch_tensor}s.

\begin{CodeChunk}
\begin{CodeInput}
R> str(materialize(rows$image))
\end{CodeInput}
\begin{CodeOutput}
List of 2
 $ :Float [1:1, 1:28, 1:28]
 $ :Float [1:1, 1:28, 1:28]
\end{CodeOutput}
\end{CodeChunk}

\begin{figure}[t]
\centering
\includegraphics[scale=0.72]{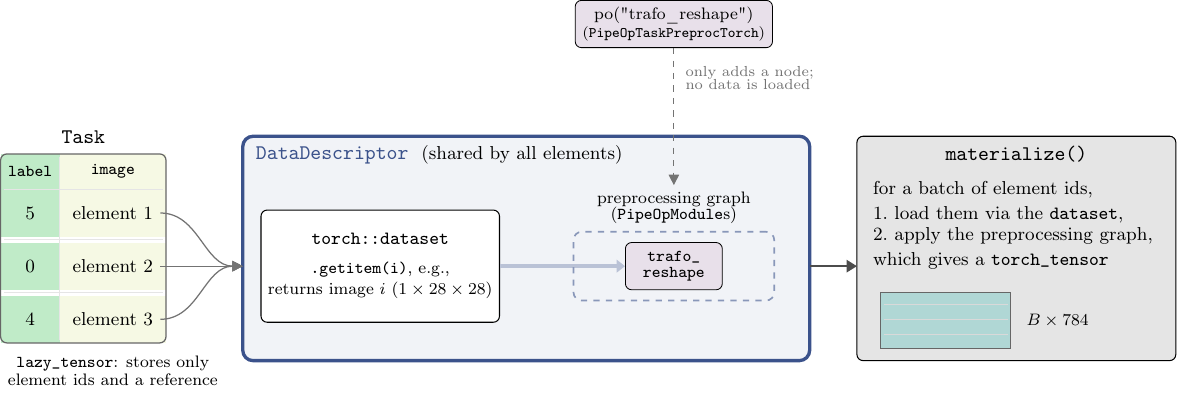}
\caption{How a \code{lazy\_tensor} works, using the MNIST task from above. The \code{image} column of the \code{Task} only stores element ids and a reference to a \code{DataDescriptor} that is shared by all elements. It contains the \code{torch::dataset} that defines how to load an element and a preprocessing graph, to which \code{PipeOpTaskPreprocTorch} operators such as \code{po("trafo\_reshape")} (see below) only add nodes without loading any data. In contrast, \code{PipeOpTaskPreproc} operators such as \code{po("scale")} transform tabular features directly. The data is only loaded and preprocessed when the tensors are materialized.}
\label{fig:lazy-tensor}
\end{figure}

\Cref{fig:lazy-tensor} illustrates how a \code{lazy_tensor} works under the hood.
A \code{lazy_tensor} can be constructed from a \code{torch::dataset} using the \code{as_lazy_tensor()} converter, which is demonstrated in~\Cref{sec:finetuning}.

\subsubsection{Data preprocessing and augmentation}\label{sec:preprocessing}

A core feature of the \code{lazy_tensor} type is that it can be transformed by \code{PipeOp}s.
Below, we scale the pixel values from $[0, 255]$ to $[0, 1]$ using \code{po("trafo_normalize")}, which subtracts \code{mean} and divides by \code{std}, and then flatten the images so they can later be fed into a simple MLP.
The two \code{PipeOp}s are chained into a \code{Graph} via the concatenation operator \code{\%\text{>}\text{>}\%} from \mlrtpipelines{}, which connects the output of its left-hand side to the input of its right-hand side.

\begin{CodeChunk}
\begin{CodeInput}
R> scale_flatten <- po("trafo_normalize", mean = 0, std = 255) %>>%
+    po("trafo_reshape", shape = c(-1, 28 * 28))
R> mnist_flat <- scale_flatten$train(mnist)[[1L]]
R> mnist_flat$head(2)
\end{CodeInput}
\begin{CodeOutput}
    label         image
   <fctr> <lazy_tensor>
1:      5   <tnsr[784]>
2:      0   <tnsr[784]>
\end{CodeOutput}
\end{CodeChunk}

Materializing the images confirms that each image has been flattened to a vector of length 784.

\begin{CodeChunk}
\begin{CodeInput}
R> str(materialize(mnist_flat$head(2)$image))
\end{CodeInput}
\begin{CodeOutput}
List of 2
 $ :Float [1:784]
 $ :Float [1:784]
\end{CodeOutput}
\end{CodeChunk}

The same transformations would also be applied when calling the \code{\$predict()} method of the \code{Graph}.
Just like loading the data, these preprocessing steps are not applied eagerly, but lazily.
They build up an internal preprocessing graph that is executed when the \code{lazy_tensor} is materialized.

Similarly, data augmentation steps are also available as \code{PipeOp}s, but their identifiers are prefixed with \code{"augment_"} instead of \code{"trafo_"}.
The full list of currently available operators can be found on the package website\footnote{\url{https://mlr3torch.mlr-org.com/articles/preprocessing_list.html}}. However, it is also relatively simple to create custom ones via \code{pipeop_preproc_torch()}.
Both the transformation and augmentation \code{PipeOp}s inherit from \code{PipeOpTaskPreprocTorch}.
The only difference between them is that preprocessing \code{PipeOp}s apply their transformation during both the train and predict stages, whereas augmentation \code{PipeOp}s are only active during the former.

\subsection[The LearnerTorch class]{The \code{LearnerTorch} class}\label{sec:learner-torch}

The \code{LearnerTorch} \pkg{R6} class is the abstract base class that inherits from the \code{mlr3::Learner} class and defines the training and prediction logic.
This logic is shared by the classification and regression learners and the generic learners for \code{TaskTorch}. Custom learning problems and extensions to other task types are described in \Cref{sec:extending-learner-task}.
Each learner in \pkg{mlr3torch} inherits from this base class and additionally defines the network architecture, the hyperparameters, supported feature types, and other metadata, as well as how to construct the \code{torch::dataset} from the input \code{Task}.
One simple example is \code{LearnerTorchMLP}, which represents a simple MLP with dropout.

\subsubsection{Configuration}

\begin{samepage} %
An \mlrttorch{} \code{Learner} can be configured via construction arguments, hyperparameters, and fields.
The standard construction arguments are:

\begin{itemize}
    \item \code{loss} :: \code{TorchLoss}, which encapsulates a \code{torch::nn_loss} and represents a loss function for training,
    \item \code{optimizer} :: \code{TorchOptimizer}, which wraps a \code{torch_optimizer} and specifies how the network parameters are learned, and
    \item \code{callbacks} :: \code{list()} of \code{TorchCallback}s, which customize the training process.
\end{itemize}
\end{samepage}

All of the above objects can themselves be configured via their own hyperparameters, and while \pkg{mlr3torch} comes with predefined, ready-to-use losses, optimizers, and callbacks, its modular design allows for easy customization, as demonstrated in~\Cref{sec:extending}.
Training can also be resumed: the checkpoint callback \code{t_clbk("checkpoint")} saves the network, the optimizer state, and the states of all other callbacks, and a subsequent training run continues from such a checkpoint via the \code{resume} hyperparameter.

Below, we construct a \code{LearnerTorchMLP} classifier which uses AdamW \citep{ref-loshchilov2017decoupled} to minimize the cross-entropy loss and maintains the training history for inspection afterwards:

\begin{CodeChunk}
\begin{CodeInput}
R> mlp <- lrn("classif.mlp",
+    loss = t_loss("cross_entropy"),
+    optimizer = t_opt("adamw", lr = 0.001),
+    callbacks = t_clbk("history"))
\end{CodeInput}
\end{CodeChunk}

The hyperparameters of the \code{Learner} are represented as a \code{ParamSet}, as defined in the \pkg{paradox} package, which offers something like a type system for hyperparameter spaces \citep{ref-paradox2024}.
It automatically verifies these types on assignment, but can also be used to sample from such spaces, which drives \mlrt{}'s tuning extensions.

The hyperparameter space of the constructed \code{Learner} is the union of the \code{ParamSet}s from the three construction arguments, some hyperparameters common to all \code{LearnerTorch} classes, and some learner-specific options that, e.g., parameterize the network architecture.
To avoid name collisions, the loss, optimizer, and callback parameters are prefixed with \code{"loss."}, \code{"opt."}, and \code{"cb.<cb-id>."}, respectively.

For the MLP, the learner-specific hyperparameters include the latent dimensions (\code{neurons}), the activation function (\code{activation}), and the dropout probability (\code{p}).
The parameter values can be set using the \code{\$set_values()} method of the \code{ParamSet}.
Below, we configure several hyperparameters, including the device, the number of epochs, the batch size, the weight decay for the optimizer, and the measures we track during training.

\begin{CodeChunk}
\begin{CodeInput}
R> mlp$param_set$set_values(
+    neurons = c(100, 200), activation = nn_relu,
+    p = 0.3, opt.weight_decay = 0.01,
+    measures_train = msr("classif.logloss"),
+    epochs = 10, batch_size = 32, device = "cpu")
\end{CodeInput}
\end{CodeChunk}

The default \code{device = "auto"} trains on a CUDA GPU if one is available; we set \code{"cpu"} here and in the following examples so that the results are reproducible also when running this code on a machine with CUDA support.

Besides construction arguments, \pkg{mlr3torch} \code{Learner}s can be configured through fields, whose main difference from hyperparameters is that they have predefined semantics that are shared between all \code{Learner}s.
These include the predict type (\code{\$predict_type}) and how to construct the validation data (\code{\$validate}).
Similarly, the \code{\$encapsulate()} method specifies how errors during training and prediction are handled, e.g., by running them in a separate process, and which \code{Learner} to use as a fallback if the primary \code{Learner} fails.
Below, we configure the \code{Learner} to make probability predictions via the \code{\$configure()} setter, which can set fields and hyperparameters simultaneously.

\begin{CodeChunk}
\begin{CodeInput}
R> mlp$configure(predict_type = "prob")
\end{CodeInput}
\end{CodeChunk}

\subsubsection{Training and prediction}\label{sec:train-predict}

The primary way to use a \code{LearnerTorch} is through its \code{\$train()} and \code{\$predict()} methods.
The former takes a \code{Task} as input and stores the trained model in the \code{\$model} field.
The \code{\$predict()} method also takes a \code{Task} as input and returns the predictions.
To allow the code in this paper to easily run without a GPU, the examples mostly use small subsets of the data and small tuning budgets.
A more realistic and thus expensive version of the code, which, among other things, trains the networks on the full datasets using a GPU, is part of the reproducibility material (\Cref{app:comp-details}).
The version shown in this article instead focuses on demonstrating the user interface.
Here, we train the MLP only on the first 1000 rows of the flattened MNIST task from above:

\begin{CodeChunk}
\begin{CodeInput}
R> mlp$train(mnist_flat, row_ids = 1:1000)
\end{CodeInput}
\end{CodeChunk}

After training, we can access the training results via the \code{\$model} field.
Most importantly, the model contains the trained network (a \code{nn_module}), the optimizer state, and the states of the callbacks.

\begin{CodeChunk}
\begin{CodeInput}
R> mlp$model$network
\end{CodeInput}
\begin{CodeOutput}
An `nn_module` containing 100,710 parameters.

-- Modules ---------------------------------------------------------------
* 0: <nn_linear> #78,500 parameters
* 1: <nn_relu> #0 parameters
* 2: <nn_dropout> #0 parameters
* 3: <nn_linear> #20,200 parameters
* 4: <nn_relu> #0 parameters
* 5: <nn_dropout> #0 parameters
* 6: <nn_linear> #2,010 parameters
\end{CodeOutput}
\end{CodeChunk}
\begin{CodeChunk}
\begin{CodeInput}
R> head(mlp$model$callbacks$history, n = 2)
\end{CodeInput}
\begin{CodeOutput}
   epoch train.classif.logloss
   <num>                 <num>
1:     1              2.073583
2:     2              1.121466
\end{CodeOutput}
\end{CodeChunk}

After training, the learner can be used to make predictions on unseen data.
Here, we evaluate the predictions using the classification error \code{Measure}.

\begin{CodeChunk}
\begin{CodeInput}
R> pred <- mlp$predict(mnist_flat, row_ids = 1001:1100)
R> pred$score(msr("classif.ce"))
\end{CodeInput}

\begin{CodeOutput}
classif.ce
      0.15
\end{CodeOutput}
\end{CodeChunk}

Instead of relying on a single train-test split, we can also conveniently use other resampling schemes for estimating the generalization error.
Below, we use 3-fold cross-validation on the first 1000 observations and then aggregate the classification error across the three folds.
\begin{CodeChunk}
\begin{CodeInput}
R> mnist_small <- mnist_flat$clone()$filter(1:1000)
R> rr <- resample(mnist_small, mlp, rsmp("cv", folds = 3))
R> rr$aggregate(msr("classif.ce"))
\end{CodeInput}

\begin{CodeOutput}
classif.ce
 0.1170182
\end{CodeOutput}
\end{CodeChunk}

\subsubsection[Learner characteristics]{\code{Learner} characteristics}\label{sec:learner-characteristics}

As with any other \code{Learner}, \code{Learner}s in \pkg{mlr3torch} are annotated with standardized metadata that describe their capabilities.
These are used internally for checking the validity of operations, but also provide valuable information to users.
The available options for these characteristics are defined in the \pkg{mlr3} base package, but they can be extended via a reflection mechanism.
They include the task type, prediction types, feature types, and properties.

\begin{itemize}
    \item The \code{task_type} defines the type of \code{Task} on which the \code{Learner} operates. \mlrttorch{} provides learners for classification, regression, and the generic \code{"torch"} task type used by \code{TaskTorch}.
    \item The \code{predict_types} indicates the types of predictions the \code{Learner} can make. The predefined \mlrttorch{} regression \code{Learner}s make response predictions, while classifiers can make both response and probability predictions. For \code{TaskTorch}, prediction formats and their encoding can be customized.
    \item The \code{feature_types} specifies which features the \code{Learner} can handle. Here, \mlrttorch{} extends the available types with the \code{lazy_tensor}.
    \item The \code{properties} are usually special abilities of a \code{Learner}, such as being able to perform feature selection, but can also be other traits. We cover the selected properties \code{"marshal"} and \code{"validation"}, as well as \code{"internal_tuning"} below.
\end{itemize}

\paragraph{Marshaling} Marshaling is the process of converting an in-memory object so that it can be (de)serialized without loss of information.
\code{Learner}s with this property require such processing before and after (de)serialization.
This is necessary for \mlrttorch{} \code{Learner}s because their weights are \code{torch_tensor}s, which are external pointers that are not natively handled by \rlang{}'s serialization mechanism.
Before serializing an \mlrttorch{} \code{Learner}, one must therefore call \code{\$marshal()}, and after deserialization call \code{\$unmarshal()}.

\begin{CodeChunk}
\begin{CodeInput}
R> pth <- tempfile()
R> mlp$marshal()
R> saveRDS(mlp, pth)
R> mlp2 <- readRDS(pth)
R> mlp2$unmarshal()
\end{CodeInput}
\end{CodeChunk}

\paragraph{Validation}

When training iterative learning procedures such as deep neural networks, one often wants to monitor the model's performance during training.
In \mlrt{}, this is possible through a standardized mechanism.
Learners with this ability have the \code{"validation"} property.
These \code{Learner}s have a \code{\$validate} field through which one configures how to create the validation \code{Task}.
One option is to set this field to the ratio of data to use for validation.

\begin{CodeChunk}
\begin{CodeInput}
R> set_validate(mlp, validate = 0.3)
\end{CodeInput}
\end{CodeChunk}

This setting is also taken into account by preprocessing \code{PipeOp}s, which apply their prediction logic to the validation data during training.

\paragraph{Internal tuning}

While \mlrt{} has extensive support for standard offline hyperparameter tuning through packages like \pkg{mlr3tuning} \citep{ref-mlr3tuning2024}, some learning algorithms can also optimize hyperparameters internally during training.
For deep neural networks, this is possible for the \code{epochs} hyperparameter and is achieved through early stopping.
Although internal tuning can be considered just another hyperparameter of a \code{Learner}, standardizing this property in the \mlrt{} base package allows for seamless combination with standard offline tuning.
\Cref{sec:tuning} demonstrates how to do so.

For in-depth coverage of the validation and internal tuning mechanisms, refer to the relevant chapter in the \pkg{mlr3} book \citep{ref-mlr3book-valid}.

\subsubsection{Parallelization}\label{sec:parallelization}
\mlrttorch{} supports parallelization at different levels.
When training on the CPU, the tensor operations of \torch{} are executed in parallel on multiple threads, whose number can be set via the hyperparameter \code{num_threads}.
By \mlrt{}'s convention, \code{num_threads} is 1, so that training does not oversubscribe the CPU when multiple models are trained in parallel via \mlrt{}'s parallelization.
Furthermore, the loading of mini-batches can be distributed across multiple background processes via the hyperparameter \code{num_workers}, which is passed on to \code{torch::dataloader()}.
This is especially useful when \code{lazy_tensor}s are loaded from disk and preprocessed, which can be slow.
Via the hyperparameter \code{device}, the network can instead be trained on a GPU, e.g., \code{"cuda"} for NVIDIA GPUs or \code{"mps"} for Apple Silicon.
While a single model cannot be trained across multiple GPUs (\Cref{sec:relation-pytorch}), several models can be trained simultaneously on different GPUs, e.g., in different resampling iterations or for different hyperparameter configurations.
Beyond single training runs, \mlrttorch{} benefits from the parallelization capabilities of \mlrt{}:
resampling, benchmarking, and tuning can be parallelized via the \pkg{future} framework \citep{ref-future}, and large experiments can be executed on high-performance computing clusters via \pkg{mlr3batchmark} \citep{ref-mlr3batchmark}.
Because \mlrttorch{} learners can be marshaled (see above), trained networks can be transferred between processes.
For more information on parallelization in \mlrt{}, we refer to \citet{ref-mlr3book-technical}.

\subsection[Neural networks as Graphs]{Neural networks as \code{Graph}s}\label{sec:graphs}

To create a new \code{nn_module} (\Cref{sec:background}), one needs to define a constructor method that initializes the network weights and submodules, as well as a \code{\$forward()} method that defines how the weights are used to transform inputs.
This allows for great flexibility, but there is little structure to the architecture.
Below, we create a simple feed-forward network with one hidden layer and ReLU activation.

\begin{CodeChunk}
\begin{CodeInput}
R> nn_simple <- nn_module("nn_simple",
+    initialize = function(d_in, d_latent, d_out) {
+      self$linear1 <- nn_linear(d_in, d_latent)
+      self$activation <- nn_relu()
+      self$linear2 <- nn_linear(d_latent, d_out)
+    },
+    forward = function(x) {
+      x <- self$linear1(x)
+      x <- self$activation(x)
+      self$linear2(x)
+    }
+  )
\end{CodeInput}
\end{CodeChunk}

The resulting \code{nn_simple} is a module generator.
Below, we create a specific instance of this architecture by providing the necessary arguments for the constructor.

\begin{CodeChunk}
\begin{CodeInput}
R> net <- nn_simple(10, 100, 1)
\end{CodeInput}
\end{CodeChunk}

We can use this neural network by simply calling it on a tensor, which internally calls the \code{\$forward()} method.

\begin{CodeChunk}
\begin{CodeInput}
R> net(torch_randn(1, 10))
\end{CodeInput}

\begin{CodeOutput}
torch_tensor
0.01 *
-9.4603
[ CPUFloatType{1,1} ][ grad_fn = <AddmmBackward0> ]
\end{CodeOutput}
\end{CodeChunk}

\subsubsection[Module Graph]{Module \code{Graph}}

While \pkg{mlr3torch} allows the construction of \code{Learner}s from generic \code{nn_module}s (see~\Cref{sec:extending-learner-task}), it also offers a systematic representation via directed acyclic graphs. This representation makes it easier, for example, to inspect or change individual components.
Such an architecture is built by assembling \code{PipeOpModule}s in an \code{mlr3pipelines::Graph}.
\Cref{tab:graph-classes} gives an overview of the classes that allow representing neural networks as \code{Graph}s.
They are introduced in the remainder of this section.
Each \code{PipeOpModule} wraps an instantiated \code{nn_module}, and the edges of the \code{Graph} define the data flow between the layers.

\begin{table}[t]
\centering
\small
\begin{tabular}{@{}l>{\raggedright\arraybackslash}p{0.52\textwidth}l@{}}
\hline
Class & Role & Category \\
\hline
\code{PipeOpModule} & Node of a module \code{Graph}; wraps an instantiated \code{nn_module} & Module graph \\
\code{nn_graph} & \code{nn_module} that runs a module \code{Graph} & Module graph \\
\code{PipeOpTorchIngress} & Entry point of a generating \code{Graph} that creates a \code{TorchIngressToken}, e.g., \code{po("torch_ingress_num")} & Generating graph \\
\code{TorchIngressToken} & Defines how the input tensor of an entry point is created from a \code{Task} & Generating graph \\
\code{PipeOpTorch} & Layer of a generating \code{Graph}, created via \code{nn()} & Generating graph \\
\code{ModelDescriptor} & Network under construction, together with the \code{Task}, the \code{TorchIngressToken}s, and the training configuration (loss, optimizer, callbacks); passed between \code{PipeOpTorch}s & Generating graph \\
\code{PipeOpTorchLoss} & Configures the loss of a \code{ModelDescriptor}, i.e., \code{po("torch_loss")} & Generating graph \\
\code{PipeOpTorchOptimizer} & Configures the optimizer of a \code{ModelDescriptor}, i.e., \code{po("torch_optimizer")} & Generating graph \\
\code{PipeOpTorchCallbacks} & Configures the callbacks of a \code{ModelDescriptor}, i.e., \code{po("torch_callbacks")} & Generating graph \\
\code{PipeOpTorchModel} & Turns the \code{ModelDescriptor} into a \code{LearnerTorch} and trains it & Generating graph \\
\hline
\end{tabular}
\caption{Classes for representing neural networks as \code{Graph}s.}
\label{tab:graph-classes}
\end{table}

Note that this uses \code{Graph} objects in an unusual way, as
\pkg{mlr3pipelines} \code{Graph}s are usually used to define preprocessing or stacking pipelines that are executed exactly once per model training run, on the entire dataset.
However, the \code{Graph}s built from \code{PipeOpModule}s encode only a single neural network forward pass and are often executed many times per training run on individual batches of samples.

Below, we define a simple feed-forward network with an input dimension of 10, a latent dimension of 100, ReLU activation, and an output layer that returns a scalar value.
Note that \code{po("module_1", ...)} is shorthand for \code{po("module", id = "module_1", ...)} and can represent an arbitrary \torch{} module that is passed via the argument \code{module}.
As in \Cref{sec:data-representation}, the \code{PipeOp}s are connected via \code{\%\text{>}\text{>}\%}.

\begin{CodeChunk}
\begin{CodeInput}
R> module_graph <- po("module_1", module = nn_linear(10, 100)) %>>%
+    po("module_2", module = nn_relu()) %>>%
+    po("module_3", module = nn_linear(100, 1))
\end{CodeInput}
\end{CodeChunk}

To perform a forward pass, we can call its \code{\$train()} method.
In this case, the method expects one input tensor and returns one output tensor.
Currently, the \code{\$predict()} phase of the \code{Graph} does nothing.

To work with this \code{Graph} using a more familiar interface, we can convert it into an \code{nn_graph}, which inherits from \code{nn_module}.
When doing so, we only need to specify the input shapes. Here, \code{NA} marks a dimension of unknown size, in this case the batch dimension.
The names of \code{shapes_in} are given by \code{module_graph\$input}.
Below, we evaluate the resulting module on two randomly generated observations.

\begin{CodeChunk}
\begin{CodeInput}
R> net <- nn_graph(module_graph,
+    shapes_in = list(module_1.input = c(NA, 10)))
R> net(torch_randn(2, 10))
\end{CodeInput}
\begin{CodeOutput}
torch_tensor
-0.1218
-0.0636
[ CPUFloatType{2,1} ][ grad_fn = <AddmmBackward0> ]
\end{CodeOutput}
\end{CodeChunk}

\subsubsection[Generating Graph]{Generating \code{Graph}}

Although it is possible to manually define module graphs, it requires specifying auxiliary parameters (such as the input dimension of a linear layer), which is both tedious and unnecessary, as they can be inferred from the shape of the input data.
Therefore, one usually creates such a module \code{Graph} through another generating \code{Graph}, which is similar in spirit to the module generator from earlier.
The generating \code{Graph} primarily consists of \code{PipeOpTorch} objects, but it can also contain other components.
The generating \code{Graph} for the above module \code{Graph} is specified below, where \code{nn("<key>")} is short for \code{po("nn_<key>", id = "<key>")}.
Note that we specify neither the input dimension for the first layer nor the output dimension of the output layer.

\begin{CodeChunk}
\begin{CodeInput}
R> graph <- po("torch_ingress_ltnsr") %>>%
+    nn("linear", out_features = 100) %>>% nn("relu") %>>% nn("head")
\end{CodeInput}
\end{CodeChunk}

While the module \code{Graph} takes in tensors, the generating \code{Graph} operates on a \code{Task} during both the training and prediction stages.
During training, the output type is a list containing a \code{ModelDescriptor} that is initialized by the \code{"torch_ingress_ltnsr"} operator.
The \code{ModelDescriptor} is the communication object between the different \code{PipeOpTorch} operators.
It contains the module \code{Graph}, the \code{Task}, the data loading instructions, as well as other metadata.
The data loading instructions are \code{TorchIngressToken}s, created by the ingress operators: each ingress operator is an entry point into the network and specifies how its input tensor is generated from the \code{Task}.
Different ingress operators exist for numeric, categorical, and \code{lazy_tensor} features; e.g., \code{po("torch_ingress_categ")} converts factors into an integer-valued tensor.
When the generating \code{Graph} is trained, the network layers (\code{nn("<key>")}) instantiate a \code{PipeOpModule} and attach it to the module \code{Graph}.
Auxiliary parameters, such as the input dimension of a linear layer, are inferred from the metadata that includes the output shape of the current position in the built-up intermediate network architecture.
Then, the \code{PipeOpTorch} computes the new output shapes and updates this and other metadata in the \code{ModelDescriptor}, and returns it.
Shapes are integer vectors whose first entry is the batch dimension, and any dimension whose size is not known in advance is \code{NA}.
Because the module \code{Graph} is a reference object, all parallel branches in the generating \code{Graph} add their modules to the same module \code{Graph}, which allows a merging operator to later join these branches.
Note that the weights of the neural network that is being built up are not yet fit to the data at this point.
Below, we train the generating \code{Graph} on the flattened MNIST \code{Task} from earlier.

\begin{CodeChunk}
\begin{CodeInput}
R> md <- graph$train(mnist_flat)[[1L]]
R> md
\end{CodeInput}
\begin{CodeOutput}
<ModelDescriptor: 4 ops>
* Ingress:  torch_ingress_ltnsr.input: [(NA,784)]
* Task:  mnist [classif]
* Callbacks:  N/A
* Optimizer:  N/A
* Loss:  N/A
* pointer:  head.output [(NA,10)]
\end{CodeOutput}
\end{CodeChunk}

If we want not only to create the neural network but also to fit its weights, we need to at least configure the optimizer and loss function in the \code{ModelDescriptor}.
We could also set some callbacks using \code{po("torch_callbacks")}.

\begin{CodeChunk}
\begin{CodeInput}
R> graph <- graph %>>%
+    po("torch_loss", t_loss("cross_entropy")) %>>%
+    po("torch_optimizer", t_opt("adamw", lr = 0.001))
\end{CodeInput}
\end{CodeChunk}

The actual training of the neural network happens via \code{PipeOpTorchModel}.
This \code{PipeOp} takes in the \code{ModelDescriptor}, converts it to a \code{LearnerTorchModel}, and trains it on the \code{Task} stored in the \code{ModelDescriptor}.
Through its hyperparameters, the \code{PipeOp} also allows specifying the generic \code{LearnerTorch} hyperparameters.

\begin{CodeChunk}
\begin{CodeInput}
R> graph <- graph %>>%
+    po("torch_model_classif", epochs = 10, batch_size = 16, device = "cpu")
\end{CodeInput}
\end{CodeChunk}

\begin{figure}
    \centering
    \includegraphics[width=\textwidth]{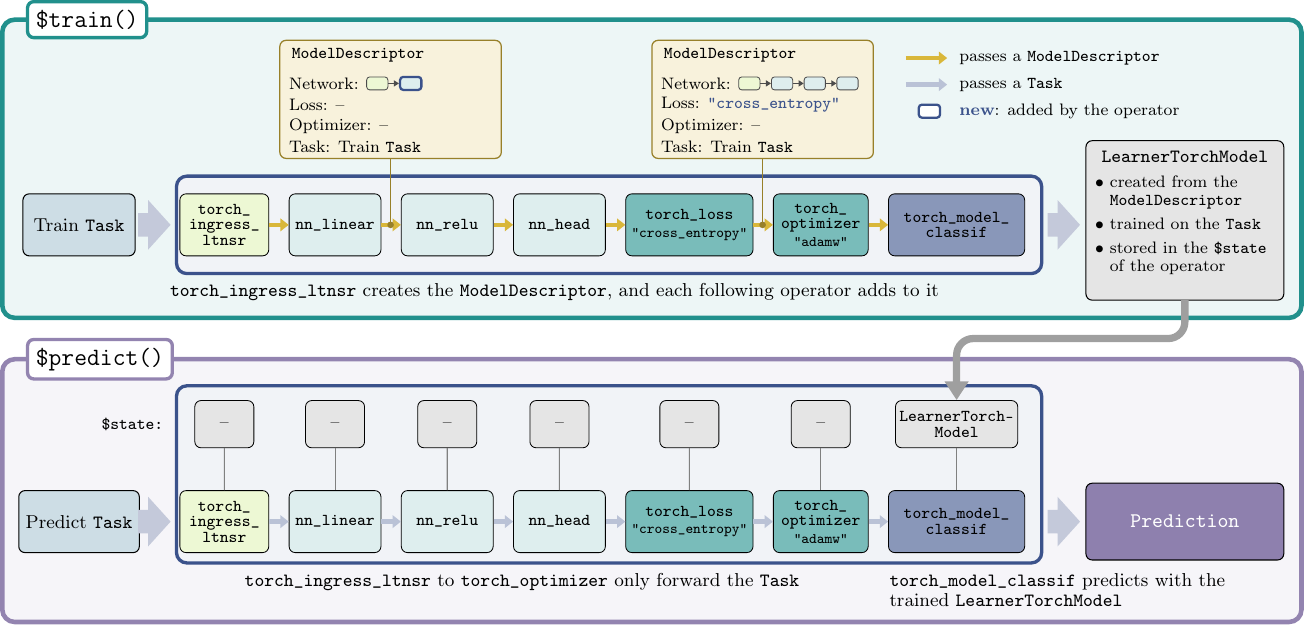}
    \caption{Training and prediction phases of a generating \code{Graph}.}
    \label{fig:pipeop-torch}
\end{figure}

During the prediction phase of the above \code{Graph}, \code{PipeOpTorch} operators mostly forward the input \code{Task}, possibly after combining multiple ones via a \code{PipeOpFeatureUnion}.
Only the \code{PipeOpTorchModel} calls the \code{\$predict()} method of the \code{LearnerTorchModel} that was stored in its state during the training phase.
This process is visualized in~\Cref{fig:pipeop-torch}.

The function \code{as_learner_torch()} converts this \code{Graph} into an \code{mlr3::Learner}, which makes it interoperable with all other components from the \pkg{mlr3} ecosystem.

\begin{CodeChunk}
\begin{CodeInput}
R> glrn <- as_learner_torch(graph)
R> glrn$train(mnist_flat, row_ids = 1:1000)
\end{CodeInput}
\end{CodeChunk}

The \code{ParamSet} of the resulting \code{Learner} contains the hyperparameters of all the individual \code{PipeOp}s that are part of its \code{Graph}.
This means that they can be changed afterwards, but more importantly, they can easily be tuned via the \pkg{mlr3tuning} extension; see~\Cref{sec:tuning}.

\subsubsection[Nonlinear Graphs]{Nonlinear \code{Graph}s}\label{sec:complex-architectures}

In this section, we show how to create more complex architectures than the single-hidden-layer MLP from earlier.
To demonstrate the construction of nonlinear graph structures, we define a residual layer.
We start by defining two independent branches for the linear and nonlinear paths, respectively.

\begin{CodeChunk}
\begin{CodeInput}
R> path_lin <- nn("linear_1")
R> path_nonlin <- nn("linear_2") %>>% nn("relu")
\end{CodeInput}
\end{CodeChunk}

The two graphs can be combined using a merge operation, in this case, summation.
The resulting graph segment is visualized in~\Cref{fig:residual-layer}.

\begin{CodeChunk}
\begin{CodeInput}
R> residual_layer <- list(path_lin, path_nonlin) %>>% nn("merge_sum")
\end{CodeInput}
\end{CodeChunk}

\subsubsection{Multi-input architectures}

It is also possible to create neural networks with more than one input (ingress).
This is necessary for defining multimodal architectures.
Here, we show how to build a tabular neural network that accepts both categorical and numeric inputs.

First, we implement two independent entry points into the network.
For the numeric path, we first select the numeric features.
This is a standard \code{PipeOp} from \mlrtpipelines{}, and it simply outputs a modified \code{Task}.
Next, \code{po("torch_ingress_num")} initializes the \code{ModelDescriptor} that represents the numeric network entry.
Subsequently, we embed the scalar values into a 10-dimensional vector using \code{nn("tokenizer_num")} \citep{ref-gorishniy2021revisiting}.
The pipeline for the categorical features is defined analogously.

\begin{CodeChunk}
\begin{CodeInput}
R> path_num <- po("select_1",
+    selector = selector_type(c("numeric", "integer"))) %>>%
+    po("torch_ingress_num") %>>%
+    nn("tokenizer_num", d_token = 10)
R> path_categ <- po("select_2", selector = selector_type("factor")) %>>%
+    po("torch_ingress_categ") %>>%
+    nn("tokenizer_categ", d_token = 10)
\end{CodeInput}
\end{CodeChunk}

\begin{figure}[t!]
    \centering
    \begin{subfigure}{0.38\textwidth}
        \centering
        \includegraphics[width=\textwidth, height = 56pt]{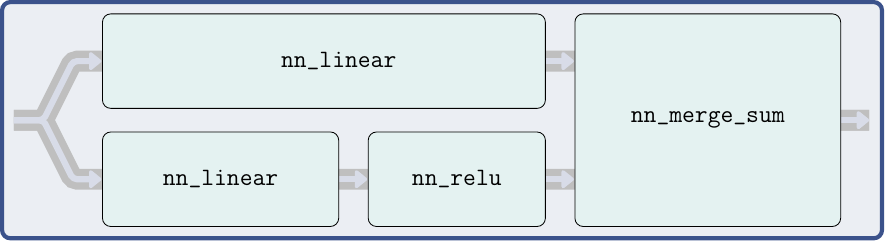}
        \caption{Residual layer.}
        \label{fig:residual-layer}
    \end{subfigure}
    \hfill
    \begin{subfigure}{0.58\textwidth}
        \centering
        \includegraphics[width=\textwidth]{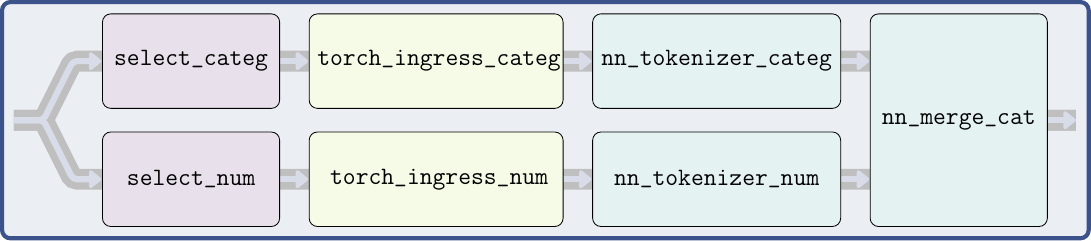}
        \caption{Neural network graph with multiple inputs.}
        \label{fig:multi-inputs}
    \end{subfigure}
    \caption{More complex network segments.}
    \label{fig:side-by-side}
\end{figure}

The two independent entry points can be combined via another merge operation, in this case by concatenation of tensors along the second dimension.
The resulting network has two inputs, where the first one expects a floating-point tensor and the second one an integer-valued tensor representing the categorical features.
The graph structures are displayed in~\Cref{fig:multi-inputs}.

\begin{CodeChunk}
\begin{CodeInput}
R> graph <- list(path_num, path_categ) %>>% nn("merge_cat", dim = 2)
\end{CodeInput}
\end{CodeChunk}

One also often wants to repeat a specific neural network segment multiple times.
This can be achieved by using the \code{PipeOpTorchBlock} operator.
This \code{PipeOp} takes in another \code{Graph} primarily consisting of \code{PipeOpTorch} objects.
During training, the \code{PipeOp} repeatedly attaches the same network segment to the neural network.

\begin{CodeChunk}
\begin{CodeInput}
R> blocks <- nn("block", residual_layer, n_blocks = 5)
\end{CodeInput}
\end{CodeChunk}

As both the number of repetitions (\code{n_blocks}) and the hyperparameters of the wrapped \code{Graph} are exposed through the \code{PipeOpTorchBlock}'s \code{ParamSet}, both can be tuned over.

\section{Extending the package}\label{sec:extending}

While the configuration options shown in the previous section already cover a lot of ground, it is also essential to be able to customize the training and prediction further.
Consequently, \pkg{mlr3torch} allows users to customize losses, optimizers, callbacks, and complete \code{Learner}s.
That these extension points also suffice for third-party packages is shown by \pkg{mlr3torchAUM} \citep{ref-mlr3torchAUM}, which adds losses for AUC optimization, optimizers, batch samplers, and performance measures for imbalanced classification that can be used directly with the \code{Learner}s of \mlrttorch{}.

\subsection{Loss and optimizer}\label{sec:extending-loss-opt}

The loss and optimizer in \mlrttorch{} are represented via the \code{TorchLoss} and \code{TorchOptimizer} classes.
They wrap a loss generator (such as \code{torch::nn_mse_loss}) and an optimizer generator (e.g., \code{torch::optim_sgd}), respectively, and annotate them with meta-information, most importantly a \code{ParamSet}.
Constructing a custom loss or optimizer, therefore, requires underlying class generators that are constructible via \code{torch::nn_module} and \code{torch::optimizer}.
The resulting constructors can be converted to their \mlrttorch{} representation via the functions \code{as_torch_loss()} and \code{as_torch_optimizer()}.
Without specifying additional arguments, all the meta-information is inferred as precisely as possible.
It is also possible to provide more specific details, such as the types of hyperparameters and their valid ranges.
Below, we create a custom \pkg{torch} loss function -- the winsorized MSE -- and convert it to its \mlrttorch{} counterpart:

\begin{CodeChunk}
\begin{CodeInput}
R> nn_winsorized_mse <- nn_module(c("nn_winsorized_mse", "nn_loss"),
+    initialize = function(max_loss) {
+      self$max_loss <- max_loss
+    },
+    forward = function(input, target) {
+      torch_mean(torch_clamp((input - target)^2, max = self$max_loss))
+    }
+  )
R> tloss <- as_torch_loss(nn_winsorized_mse)
R> tloss
\end{CodeInput}
\begin{CodeOutput}
<TorchLoss:nn_winsorized_mse> nn_winsorized_mse
* Generator: nn_winsorized_mse
* Parameters: list()
* Packages: torch,mlr3torch
* Task Types: classif,regr,torch
\end{CodeOutput}
\end{CodeChunk}

This object could now be passed as the construction argument \code{loss} when initializing a new \code{LearnerTorch}.

\subsection{Callbacks}\label{sec:extending-callbacks}

The callback mechanism in \mlrttorch{} allows hooking into the training loop without having to write everything from scratch.
It is implemented via the \pkg{R6} class \code{TorchCallback}, which wraps the \code{CallbackSet} class.
Analogously to the optimizer and loss, the latter can be converted to the former via \code{as_torch_callback()}.
The design of the callback mechanism is similar to how it is implemented in frameworks such as \keras{} or \luz{} \citep{ref-chollet2018keras, ref-luz2025}.

A \code{CallbackSet} is a collection of hooks, which are invoked at specific points in the training process.
The name of the method in the callback defines at which time in the training process it is called, such as \code{on_after_backward}.
Information from the training loop is communicated to the callback via a \code{ContextTorch}, which stores the network, the current training batch, and many other objects.
Different callback hooks from a \code{CallbackSet} can communicate with each other by storing information in the \code{CallbackSet} itself.
A callback can also store information in the \code{Learner} after training by implementing the \code{\$state_dict()} and \code{\$load_state_dict()} methods.
These states are also saved by the checkpoint callback, so that they are restored when training is resumed (\Cref{sec:learner-torch}).
Further predefined callbacks support, e.g., experiment tracking: \code{t_clbk("tb")} logs the training loss as well as the training and validation metrics to \pkg{TensorBoard} via the \pkg{tfevents} package \citep{ref-tfevents}.

A \code{TorchCallback} object can be created via \code{torch_callback()}.
We demonstrate this below by implementing gradient clipping, which has two hyperparameters representing the maximally allowed norm value and the type of norm.
Besides the methods for its state, it only implements the hook that is run after the backward call.
After training, it writes the history of norms into the state of the \code{Learner} for inspection.

\begin{CodeChunk}
\begin{CodeInput}
R> gradient_clipper <- torch_callback("gradient_clipper",
+    initialize = function(max_norm, norm_type) {
+      self$norms <- numeric()
+      self$max_norm <- max_norm
+      self$norm_type <- norm_type
+    },
+    on_after_backward = function() {
+      norm <- nn_utils_clip_grad_norm_(self$ctx$network$parameters,
+        self$max_norm, self$norm_type)
+      self$norms <- c(self$norms, norm$item())
+    },
+    state_dict = function() {
+      self$norms
+    },
+    load_state_dict = function(state_dict) {
+      self$norms <- state_dict
+    }
+  )
\end{CodeInput}
\end{CodeChunk}

This object could now be passed to a \code{LearnerTorch} during construction and would limit the norms of the gradients to the specified upper bound.

\subsection[Learners and task types]{\code{Learner}s and task types}\label{sec:extending-learner-task}

When the construction of custom architectures, as shown in \Cref{sec:graphs}, is not enough, it is also possible to inherit from \code{LearnerTorch}.
While it is in principle possible to override the whole training and prediction logic (private methods \code{\$.train()} and \code{\$.predict()}), it is usually sufficient to override one or more of the following methods:

\begin{itemize}
    \item \code{\$initialize(id, loss, optimizer, callbacks, ...)}, the constructor, which defines learner-specific hyperparameters and other metadata, such as the supported feature types or the task type.
    \item \code{\$.network(task, param_vals)}, which is the network constructor and must return an \code{nn_module} for the specified hyperparameters and input \code{Task}.
    \item \code{\$.ingress_tokens(task, param_vals)}, which defines how the task's features are converted into network inputs by returning a named list of \code{TorchIngressToken}s.
    \item \code{\$.dataset(task, param_vals)}, which provides an alternative for more control over data construction. It constructs a \code{torch::dataset} for a given task.
    \item \code{\$.encode_prediction(network_output, task)}, which converts the network output into a list suitable for constructing an \code{mlr3::Prediction} object.
\end{itemize}

The help page of \code{LearnerTorch} describes these interfaces in more detail.

If one instead defines the neural network architecture as a \code{Graph} and is missing a specific layer, one can add a new \code{PipeOpTorch} via the \code{pipeop_torch()} construction function.
\Cref{sec:multimodal} shows how to use this function.
The package website contains an article with more details on how to do this\footnote{\url{https://mlr3torch.mlr-org.com/articles/custom_pipeop_torch.html}}.

\subsubsection[Custom Learners for classification and regression]{Custom \code{Learner}s for classification and regression}\label{sec:extending-learner}

If one only wants to customize the neural network architecture, it is usually not necessary to create a custom class.
Instead, one can use the \code{LearnerTorchModule} class, which allows exactly this and has the following additional construction arguments:
\begin{itemize}
    \item \code{module_generator}, which must be an \code{nn_module_generator} that takes as construction argument a \code{task} and possibly other parameters.
    \item \code{ingress_tokens}, which is a named list of \code{TorchIngressToken}s that define how the data is loaded from the input \code{Task}. The names must correspond to the arguments of the \code{module_generator}'s \code{\$forward()} method.
\end{itemize}

We showcase this using a simple MLP with two architecture parameters, namely the latent dimension and the network depth.
Note that all auxiliary parameters are inferred from the \code{task}.
In particular, the number of output units is determined by \code{output_dim_for()}, which returns the output dimension a network needs for a given task: a single unit for regression and binary classification, and one unit per class for multiclass classification.

\begin{CodeChunk}
\begin{CodeInput}
R> nn_ffn <- nn_module("nn_ffn",
+    initialize = function(task, latent_dim, n_layers) {
+      dims <- c(task$n_features, rep(latent_dim, n_layers),
+        output_dim_for(task))
+      modules <- unlist(lapply(seq_len(length(dims) - 1), function(i) {
+        if (i < length(dims) - 1) {
+          list(nn_linear(dims[i], dims[i + 1]), nn_relu())
+        } else {
+          list(nn_linear(dims[i], dims[i + 1]))
+        }
+      }), recursive = FALSE)
+      self$network <- do.call(nn_sequential, modules)
+    },
+    forward = function(x) self$network(x)
+  )
\end{CodeInput}
\end{CodeChunk}

Next, we need to define how a \code{Task} is converted to \code{torch_tensor}s during training and prediction.
In our case, the model loads all numeric columns and concatenates them into a single tensor that is passed as the argument \code{x} to the network.

\begin{CodeChunk}
\begin{CodeInput}
R> num_input <- list(x = ingress_num())
R> num_input
\end{CodeInput}

\begin{CodeOutput}
$x
Ingress: Task[selector_type(c("numeric", "integer"))] --> Tensor()
\end{CodeOutput}
\end{CodeChunk}

Next, we create the \code{LearnerTorchModule} and set the network's hyperparameters, which are part of the \code{Learner}'s \code{ParamSet}.

\begin{CodeChunk}
\begin{CodeInput}
R> lrn_ffn <- lrn("classif.module",
+    module_generator = nn_ffn,
+    ingress_tokens = num_input,
+    latent_dim = 100, n_layers = 5)
\end{CodeInput}
\end{CodeChunk}

We could now proceed and train this object on an \mlrt{} classification \code{Task} with numeric features.

\subsubsection{Other learning problems}\label{sec:other-task-types}

While regression and classification are supported out of the box in \mlrttorch{}, custom learning problems can also be represented via the generic \code{TaskTorch} class.
When creating such a task via \code{as_task_torch()}, one can specify its default performance measure, the default prediction encoder (how the network outputs are converted to an \code{mlr3::Prediction} object), as well as a network's expected output dimension.
Performance measures for such a custom learning problem are represented via the generic \code{MeasureTorch} class, which can be constructed using \code{msr_torch()}.
Creating a \code{LearnerTorch} for such a modeling problem requires specifying the architecture and data loading.
For a hands-on tutorial on how to do this, see the Custom Learning Problems article on the package website\footnote{\url{https://mlr3torch.mlr-org.com/articles/custom_task_type.html}}, which covers a multi-label classification problem and an unsupervised autoencoder.

While the flexibility of these generic classes is convenient, it comes at the cost of having fewer compatibility and correctness checks and places more responsibility on the user.
If the goal is instead to extend \mlrttorch{} to new task types, two cases can be distinguished.
If the task type already exists in the \mlrt{} ecosystem, such as survival analysis in \pkg{mlr3proba} \citep{sonabend2021mlr3proba} its \code{Task}, \code{Prediction}, and \code{Measure} classes can be used and only \mlrttorch{} support needs to be added.
Otherwise, a new task type first has to be registered in \mlrt{} with dedicated \code{Task}, \code{Learner}, \code{Prediction}, and \code{Measure} classes.
In both cases, \mlrttorch{} then supports the task type once methods for a few generics are implemented, most importantly \code{output_dim_for()} for the output dimension of the network, \code{get_target_batchgetter()} for loading the target, and \code{encode_prediction()} for converting the network output into a prediction.
The article Adding a New Task Type on the package website\footnote{\url{https://mlr3torch.mlr-org.com/articles/new_task_type.html}} documents how to properly extend \mlrttorch{} to new task types in third-party packages.

\section{Use cases}\label{sec:use-cases}

We now demonstrate some of \pkg{mlr3torch}'s features by tuning a neural network on a tabular regression problem (\Cref{sec:tuning}), fine-tuning a pretrained image network from \pkg{torchvision} (\Cref{sec:finetuning}), and constructing a network architecture for a problem that contains both tabular features and images (\Cref{sec:multimodal}).
Note that the examples are chosen for simplicity and are intended to illustrate the functionality of \mlrttorch{}.
The architectures, configurations, and datasets are chosen so the examples can quickly be reproduced on a CPU.
In particular, the \code{"mtcars"} task used in \Cref{sec:tuning} has only 32 observations.
The more expensive version of the code (\Cref{app:comp-details}) therefore uses a larger dataset for the neural architecture search.

\subsection{Simple neural architecture search}\label{sec:tuning}

In this section, we construct an attention-based tabular neural network and simultaneously tune its architecture, optimizer, and number of epochs.
We use the predefined \code{"mtcars"} example problem \citep{ref-mtcars}.

\begin{CodeChunk}
\begin{CodeInput}
R> task <- tsk("mtcars")
R> task
\end{CodeInput}
\begin{CodeOutput}
-- <TaskRegr> (32x11): Motor Trends --------------------------------------
* Target: mpg
* Properties: -
* Features (10):
  * dbl (10): am, carb, cyl, disp, drat, gear, hp, qsec, vs, wt
\end{CodeOutput}
\end{CodeChunk}

\subsubsection{Defining the learning pipeline}
The input of our network architecture is a single numeric tensor, so the first step in the learning pipeline is the initialization of a numeric network entry via \code{PipeOpTorchIngressNumeric}.

\begin{CodeChunk}
\begin{CodeInput}
R> ingress <- po("torch_ingress_num")
\end{CodeInput}
\end{CodeChunk}

The subsequent \code{PipeOp}s define the layers of the neural network.
Instead of directly passing the feature vector to an MLP, we build a simple attention architecture.
Its first step is a feature tokenizer, which embeds every numeric feature into its own vector of dimension 8 using a linear layer.

\begin{CodeChunk}
\begin{CodeInput}
R> embedding <- nn("tokenizer_num", d_token = 8)
\end{CodeInput}
\end{CodeChunk}

On these feature tokens, we then repeatedly apply a block that consists of multi-head self-attention \citep{vaswani2017attention}, layer normalization \citep{ba2016layernormalization}, and a small MLP.
The attention operates across the feature tokens, allowing the network to model interactions between them.
The subsequent MLP is applied to each token separately and consists of a linear transformation followed by an activation function and a dropout layer.
We can allow different choices of activation functions via the \code{ppl("branch")} construct, which has a \code{branch.selection} parameter.
In the example below, we allow for two choices, namely a ReLU and a sigmoid activation function.

\begin{CodeChunk}
\begin{CodeInput}
R> block <- nn("multihead_attention", num_heads = 2) %>>%
+    nn("layer_norm", dims = 1) %>>%
+    nn("linear", out_features = 16) %>>%
+    ppl("branch", list(relu = nn("relu"), sigmoid = nn("sigmoid"))) %>>%
+    nn("dropout")
\end{CodeInput}
\end{CodeChunk}

This also demonstrates one of the strengths of \pkg{mlr3torch}: because \code{PipeOpTorch} operators are just \code{PipeOp}s, they can be combined with other operators from \pkg{mlr3pipelines} as long as their input and output types are compatible.

To repeat the above-defined segment one or more times, we can use the \code{PipeOpTorchBlock} class, which we have already covered earlier.
Afterwards, the resulting tokens are flattened into a single vector before they are passed to the output head.

\begin{CodeChunk}
\begin{CodeInput}
R> architecture <- embedding %>>% nn("block", block) %>>%
+    nn("flatten") %>>% nn("head")
\end{CodeInput}
\end{CodeChunk}

Next, we set the loss function to MSE and the optimizer to AdamW.

\begin{CodeChunk}
\begin{CodeInput}
R> config <- po("torch_loss", loss = t_loss("mse")) %>>%
+    po("torch_optimizer", optimizer = t_opt("adamw"))
\end{CodeInput}
\end{CodeChunk}

The final step is the \code{PipeOpTorchModelRegr}, which takes in a \code{ModelDescriptor} and does the actual training:

\begin{CodeChunk}
\begin{CodeInput}
R> model <- po("torch_model_regr", device = "cpu", batch_size = 32)
\end{CodeInput}
\end{CodeChunk}

Next, we combine all pieces into a single learning pipeline and convert it to a \code{Learner}.
Because the features are on very different scales, we standardize them via \code{po("scale")}.

\begin{CodeChunk}
\begin{CodeInput}
R> pipeline <- po("scale") %>>% ingress %>>%
+    architecture %>>% config %>>% model
R> learner <- as_learner_torch(pipeline)
R> learner$id <- "attention_nn"
\end{CodeInput}
\end{CodeChunk}

The \code{ParamSet} of the \code{GraphLearnerTorch} is the union of the \code{ParamSet}s of the individual \code{PipeOp}s, over which we now tune.

\subsubsection{Defining the search space}
In order to tune our \code{Learner}, we need to define the search space.
We tune the embedding dimension over $\{4, 8, 16\}$, the number of attention heads over $\{1, 2, 4\}$, the attention dropout in $[0, 0.5]$, and the latent dimension of the block's MLP over $\{8, 16, 32\}$.
Furthermore, we repeat the block between 1 and 3 times, use ReLU or sigmoid, set the dropout ratio to a value in $[0.1, 0.9]$, and search the optimizer's learning rate in the interval $[10^{-4}, 10^{-1}]$ on a logarithmic scale.

\begin{CodeChunk}
\begin{CodeInput}
R> library("mlr3tuning")
R> learner$param_set$set_values(
+    tokenizer_num.d_token = to_tune(c(4, 8, 16)),
+    block.multihead_attention.num_heads = to_tune(c(1, 2, 4)),
+    block.multihead_attention.dropout = to_tune(0, 0.5),
+    block.linear.out_features = to_tune(c(8, 16, 32)),
+    block.n_blocks = to_tune(1, 3),
+    block.branch.selection = to_tune(c("relu", "sigmoid")),
+    block.dropout.p = to_tune(0.1, 0.9),
+    torch_optimizer.lr = to_tune(10^-4, 10^-1, logscale = TRUE))
\end{CodeInput}
\end{CodeChunk}

We tune all the above hyperparameters in the standard offline way using random search.
In contrast, we tune the \code{epochs} hyperparameter using the learner-internal early stopping procedure that was covered in \Cref{sec:learner-torch}.
By setting the \code{\$validate} field to \code{"test"}, we stop early on the test set of the resampling used for the offline hyperparameter tuning, i.e., the validation data of the tuning process.

\begin{CodeChunk}
\begin{CodeInput}
R> set_validate(learner, "test")
\end{CodeInput}
\end{CodeChunk}

Finally, we set the patience parameter for the early stopping, the maximum number of epochs, and the validation measure.
By setting \code{internal = TRUE} for the epochs, we enable tuning of the epochs via early stopping.

\begin{CodeChunk}
\begin{CodeInput}
R> learner$param_set$set_values(
+    torch_model_regr.patience = 5,
+    torch_model_regr.measures_valid = msr("regr.mse"),
+    torch_model_regr.epochs = to_tune(upper = 100, internal = TRUE))
\end{CodeInput}
\end{CodeChunk}

\subsubsection{Running the tuning procedure}

As the tuning method, we use random search via the \pkg{mlr3tuning} extension \citep{ref-mlr3tuning2024} for simplicity, but more complex optimization routines such as Bayesian optimization are also supported via the \pkg{mlr3mbo} extension \citep{ref-mlr3mbo, ref-bischl2023hyperparameter}.
For the evaluation of the hyperparameters, we select holdout resampling (for demonstrative purposes) and use the best validation score -- which we configured via the \code{measures_valid} hyperparameter -- as the tuning measure.
We run the tuning loop for 20 iterations.

\begin{CodeChunk}
\begin{CodeInput}
R> ti <- tune(
+    tuner = tnr("random_search"),
+    resampling = rsmp("holdout"),
+    measure = msr("best_valid_score", minimize = TRUE),
+    learner = learner,
+    term_evals = 20,
+    task = task)
R> tuned <- learner$param_set$search_space()$ids()
R> pvals <- lapply(ti$result_learner_param_vals[tuned],
+    function(x) if (is.numeric(x)) signif(x, 3) else x)
R> cat(paste("*", names(pvals), "=", pvals,
+    collapse = "\n"), "\n")
\end{CodeInput}
\begin{CodeOutput}
* tokenizer_num.d_token = 16
* block.n_blocks = 1
* block.multihead_attention.num_heads = 4
* block.multihead_attention.dropout = 0.246
* block.linear.out_features = 32
* block.branch.selection = sigmoid
* block.dropout.p = 0.711
* torch_optimizer.lr = 0.0435
* torch_model_regr.epochs = 16
\end{CodeOutput}
\end{CodeChunk}

We could now re-fit our network on the whole data with the configuration found above to obtain our final model.

\subsection{Fine-tuning an image network}\label{sec:finetuning}\nopagebreak

Fine-tuning is a common approach in DL that, instead of starting with randomly initialized weights, starts with pretrained networks and adjusts them to new data.
This often allows the model to transfer some of the learned representations from the pretraining to the new task at hand.
This both reduces training time and improves performance, especially if task-specific data is limited.

In this section, we show how \pkg{mlr3torch} allows fine-tuning a pretrained image classification network from the \pkg{torchvision} package.
We fine-tune the ConvNeXt-T model \citep{ref-liu2022convnet} on the Dogs vs.\ Cats dataset \citep{ref-dogs-vs-cats2013}, which is a binary image classification problem.

\subsubsection[Task definition]{\code{Task} definition}

As a first step, we have to construct the classification \code{Task}, the data for which we access via the \pkg{torchdatasets} package \citep{torchdatasets}.

\begin{CodeChunk}
\begin{CodeInput}
R> library("torchdatasets")
R> dvc <- dogs_vs_cats_dataset("data", download = TRUE)
\end{CodeInput}
\end{CodeChunk}

Next, we create a \code{lazy_tensor} vector for the images.
To do so, we create a custom \code{torch::dataset} class that stores the image paths and loads them from disk.
Note that \pkg{mlr3torch} requires the \code{\$.getitem()} method to return a named list of \code{torch_tensor}s.
During data retrieval, we also move the last (channel) dimension of the tensor to the front, as \pkg{torchvision} models expect the RGB channels to be the first dimension.

\begin{CodeChunk}
\begin{CodeInput}
R> ds <- torch::dataset("dogs_vs_cats",
+    initialize = function(pths) {
+      self$pths <- pths
+    },
+    .getitem = function(i) {
+      image <- torchvision::base_loader(self$pths[i])
+      list(image = torch_tensor(image)$permute(c(3, 1, 2)))
+    },
+    .length = function() {
+      length(self$pths)
+    }
+  )
\end{CodeInput}
\end{CodeChunk}

In order to initialize this dataset, we need to create a vector containing the image paths.

\begin{CodeChunk}
\begin{CodeInput}
R> paths <- list.files(file.path("data", "dogs-vs-cats/train"),
+    full.names = TRUE)
R> dogs_vs_cats <- ds(paths)
\end{CodeInput}
\end{CodeChunk}

To convert this into a \code{lazy_tensor}, we can use the \code{as_lazy_tensor()} converter.
Here we specify the output shapes of the images to be \code{NULL}, which denotes that the items have varying shapes.

\begin{CodeChunk}
\begin{CodeInput}
R> lt <- as_lazy_tensor(dogs_vs_cats, list(image = NULL))
\end{CodeInput}
\end{CodeChunk}

Next, we construct the target vector.
In this example, this information is part of the file names.

\begin{CodeChunk}
\begin{CodeInput}
R> labels <- ifelse(grepl("dog\\.\\d+\\.jpg", paths), "dog", "cat")
R> table(labels)
\end{CodeInput}
\begin{CodeOutput}
labels
  cat   dog
12500 12500
\end{CodeOutput}
\end{CodeChunk}

Next, we construct a \code{data.table} \citep{ref-datatable2024} that contains the feature and target columns and convert it to an \code{mlr3::Task}.

\begin{CodeChunk}
\begin{CodeInput}
R> tbl <- data.table(image = lt, class = labels)
R> task <- as_task_classif(tbl, target = "class", id = "dogs_vs_cats")
R> task
\end{CodeInput}
\begin{CodeOutput}
-- <TaskClassif> (25000x2) -----------------------------------------------
* Target: class
* Properties: twoclass
* Features (1):
  * lt (1): image
* Target classes: cat (positive class, 50%), dog (50%)
\end{CodeOutput}
\end{CodeChunk}

\subsubsection{Learning pipeline}

Besides fine-tuning, we also integrate data augmentation and preprocessing \code{PipeOp}s into our learning pipeline.
Below, we define the data augmentation.

\begin{CodeChunk}
\begin{CodeInput}
R> augment <- po("augment_random_horizontal_flip", p = 0.5)
\end{CodeInput}
\end{CodeChunk}

For preprocessing, we resize each image to size $(3, 224, 224)$, which is the resolution on which the pretrained ConvNeXt-T weights were trained.
By default, this is done using bilinear interpolation.

\begin{CodeChunk}
\begin{CodeInput}
R> preprocess <- po("trafo_resize", size = c(224, 224))
\end{CodeInput}
\end{CodeChunk}

Most \pkg{torchvision} models also expect their inputs to be normalized, so we normalize each channel using the mean and standard deviation of the ImageNet training data, on which the weights were pretrained.

\begin{CodeChunk}
\begin{CodeInput}
R> normalize <- po("trafo_normalize",
+    mean = c(0.485, 0.456, 0.406), std = c(0.229, 0.224, 0.225))
\end{CodeInput}
\end{CodeChunk}

For better training performance, we freeze all but the head of the network for the first two epochs and only then train the remaining parameters.
We achieve this via the \code{"unfreeze"} callback and only need to specify which parameters to train from the beginning and when to unfreeze the other weights.
The names of the weights (e.g., \code{"head.weight"}) are defined by the underlying \code{nn_module} and can, for example, be looked up in its source code\footnote{\url{https://github.com/mlverse/torchvision/blob/main/R/models-convnext.R}}.
In general, the help page of a predefined \code{Learner} links to the implementation of its architecture if it comes from another package such as \pkg{torchvision}.
Otherwise, consult the source code of \mlrttorch{} itself.

\begin{CodeChunk}
\begin{CodeInput}
R> unfreezer <- t_clbk("unfreeze",
+    starting_weights = select_name(c("head.weight", "head.bias")),
+    unfreeze = data.table(epoch = 3, weights = select_all()))
\end{CodeInput}
\end{CodeChunk}

The final step is the construction of the ConvNeXt-T, which has a parameter \code{pretrained} that determines whether to use the pretrained weights.

\begin{CodeChunk}
\begin{CodeInput}
R> convnext <- lrn("classif.convnext_tiny_1k",
+    pretrained = TRUE, epochs = 5, device = "cpu", batch_size = 32,
+    opt.lr = 1e-4, measures_valid = msr("classif.acc"),
+    callbacks = list(unfreezer, t_clbk("history")))
\end{CodeInput}
\end{CodeChunk}

When the \code{Learner} is instructed to use the pretrained weights, \pkg{torchvision} downloads (and caches) those weights upon first use, and the \mlrttorch{} \code{Learner} replaces the output layer of the network with a new head with the correct output dimension for the problem at hand.

\subsubsection{Training}

Now, we combine these steps into a single new \code{GraphLearnerTorch}, configure its validation data, train it, and print the validation accuracy.
We conduct the training on only a subset of the data -- 100 rows from each class -- to make the code run quickly.
Due to the use of pretrained weights, the model quickly reaches a good performance despite being trained on only a few examples.
Note that the uncertainty of the validation performance is rather high as we only use 1/3 of 200 observations.

\begin{CodeChunk}
\begin{CodeInput}
R> learner <- as_learner_torch(
+    augment %>>% preprocess %>>% normalize %>>% convnext)
R> learner$id <- "convnext"
R> set_validate(learner, 1 / 3)
R> learner$train(task, 12401:12600)
R> learner$model$classif.convnext_tiny_1k$model$callbacks$history
\end{CodeInput}
\begin{CodeOutput}
   epoch valid.classif.acc
   <num>             <num>
1:     1         0.3582090
2:     2         0.3731343
3:     3         0.9552239
4:     4         1.0000000
5:     5         1.0000000
\end{CodeOutput}
\end{CodeChunk}

\subsection{Multimodal data}\label{sec:multimodal}

In this section, we define a network architecture for the \code{"melanoma"} example task, where the goal is to predict whether a tumor is malignant or benign \citep{ref-rotemberg2021melanoma}.
The dataset contains both images and three tabular features: the approximate age and sex of the patient, and the anatomical site of the lesion.
Furthermore, the data is grouped by \code{patient_id}, which \pkg{mlr3} automatically respects during resampling.

\begin{CodeChunk}
\begin{CodeInput}
R> task <- tsk("melanoma")
R> task
\end{CodeInput}

\begin{CodeOutput}
-- <TaskClassif> (32701x5): Melanoma Classification ----------------------
* Target: outcome
* Properties: twoclass, groups
* Features (4):
  * fct (2): anatom_site_general_challenge, sex
  * int (1): age_approx
  * lt (1): image
* Groups: patient_id
* Target classes: malignant (positive class, 2%), benign (98%)
\end{CodeOutput}
\end{CodeChunk}

Because malignant tumors are much rarer than benign ones, the class distribution is imbalanced.

\begin{CodeChunk}
\begin{CodeInput}
R> table(task$truth())
\end{CodeInput}
\begin{CodeOutput}
malignant    benign
      581     32120
\end{CodeOutput}
\end{CodeChunk}

Also, for some observations, the approximate age is missing.

\begin{CodeChunk}
\begin{CodeInput}
R> task$missings("age_approx")
\end{CodeInput}
\begin{CodeOutput}
age_approx
        44
\end{CodeOutput}
\end{CodeChunk}

The network architecture we create for this dataset represents every (one-hot encoded) tabular column and every image patch as a token and lets a transformer process all of them jointly.
For the tabular features, we first select them, sample the missing values of the age from their empirical distribution, and encode the categorical features one-hot.
Then, we mark the entry point of the neural network and embed each of these columns into its own token of dimension 64 via a numeric tokenizer.

\begin{CodeChunk}
\begin{CodeInput}
R> path_tabular <- po("select_1",
+      selector = selector_type(c("integer", "factor"))) %>>%
+    po("imputehist") %>>%
+    po("encode", method = "one-hot") %>>%
+    po("torch_ingress_num") %>>%
+    nn("tokenizer_num", d_token = 64)
\end{CodeInput}
\end{CodeChunk}

For the image column, less preprocessing is required, and we process the image with attention only, following the vision transformer \citep{dosovitskiy2021image}.
The $128 \times 128$ image is split into a grid of $8 \times 8$ disjoint patches of size $16 \times 16$ that are flattened.
For this, we use \code{nn("fn")}, which allows using an arbitrary tensor function in a network graph, together with \code{torch::nnf_unfold()}, which extracts the flattened patches as the columns of a tensor.
Each of these 64 patches is embedded using a linear layer with output dimension 64.

\begin{CodeChunk}
\begin{CodeInput}
R> patchify <- function(x) {
+    nnf_unfold(x, kernel_size = 16, stride = 16)$transpose(2, 3)
+  }
R> path_image <- po("select_2", selector = selector_name("image")) %>>%
+    po("torch_ingress_ltnsr", shape = c(NA, 3, 128, 128)) %>>%
+    nn("fn_1", fn = patchify) %>>%
+    nn("linear", out_features = 64)
\end{CodeInput}
\end{CodeChunk}

Because self-attention does not take the order of the tokens into account, we add a learnable positional embedding: one for each image patch.
Because no predefined \code{PipeOpTorch} exists for this, we create a custom one using \code{pipeop_torch()}.
It works similarly to \code{nn_module()} from \pkg{torch} but additionally requires implementing the shape inference function, which is trivial in this case.
In general, \code{shapes_out()} receives the list of input shapes and returns the list of output shapes as integer vectors. Because any input dimension may be \code{NA}, the function must not assume that a dimension it reads is known, and unknown output dimensions are again \code{NA}.

\begin{CodeChunk}
\begin{CodeInput}
R> po_torch_pos_embedding <- pipeop_torch("nn_pos_embedding",
+    initialize = function(shapes_in) {
+      self$weight <- nn_parameter(torch_randn(shapes_in[[1]][-1]) * 0.02)
+    },
+    forward = function(input) input + self$weight,
+    shapes_out = function(shapes_in) shapes_in
+  )
R> path_image <- path_image %>>% po_torch_pos_embedding$new()
\end{CodeInput}
\end{CodeChunk}

Next, we concatenate the tabular and image tokens and process the resulting sequence with two transformer encoder layers \citep{vaswani2017attention}.
Afterwards, we average the output tokens and pass the result to the classification head.

\begin{CodeChunk}
\begin{CodeInput}
R> architecture <- list(path_tabular, path_image) %>>%
+    nn("merge_cat", dim = 2) %>>%
+    nn("block", nn("transformer_encoder_layer", nhead = 4,
+      dim_feedforward = 128), n_blocks = 2) %>>%
+    nn("fn_2", fn = function(x) x$mean(dim = 2)) %>>%
+    nn("head")
\end{CodeInput}
\end{CodeChunk}

Finally, we set the loss function, optimizer, and training arguments.
Because the problem is highly imbalanced, we increase the weight of the positive class by a factor of 10.
For the optimizer, we again use AdamW, and we train for 4 epochs with a batch size of 32.

\begin{CodeChunk}
\begin{CodeInput}
R> model <- architecture %>>%
+    po("torch_loss",
+      t_loss("cross_entropy", class_weight = torch_tensor(10))) %>>%
+    po("torch_optimizer", t_opt("adamw", lr = 0.0005)) %>>%
+    po("torch_model_classif", epochs = 4, batch_size = 32, device = "cpu",
+      predict_type = "prob")
\end{CodeInput}
\end{CodeChunk}

As further measures to address the imbalanced class distribution, we upsample the minority class by a factor of 4 and use various data augmentation methods, namely random flips and random crops of the images.

\begin{CodeChunk}
\begin{CodeInput}
R> preprocessing <- po("classbalancing", ratio = 4, reference = "minor",
+      adjust = "minor") %>>%
+    po("augment_random_horizontal_flip") %>>%
+    po("augment_random_vertical_flip") %>>%
+    po("augment_random_crop", size = c(128, 128), padding = 8)
R> glrn <- as_learner_torch(preprocessing %>>% model)
\end{CodeInput}
\end{CodeChunk}

\begin{figure}[t!]
    \centering
    \includegraphics[width=0.55\textwidth]{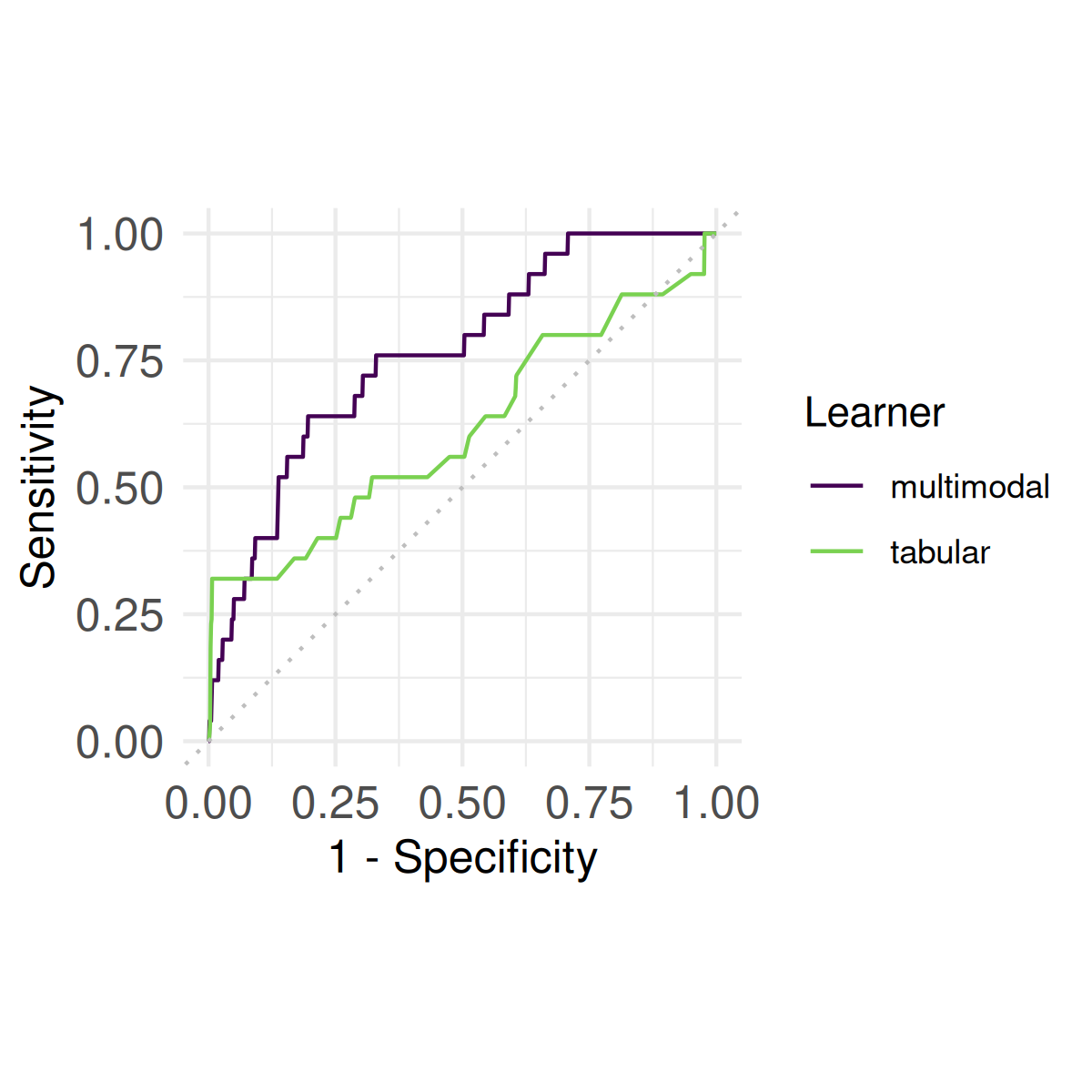}
    \caption{ROC curves of the multimodal neural network and the logistic regression baseline that only uses the tabular features, evaluated on the same holdout split.}
    \label{fig:roc-curve}
\end{figure}

Finally, we compare the multimodal network with a baseline that only uses the tabular features, namely a logistic regression from the \pkg{mlr3learners} package \citep{ref-mlr3learners}, for which we use the same imputation as above.
With \code{benchmark_grid()}, we create a design in which both \code{Learner}s are evaluated on the same holdout split, which respects the grouping variable \code{patient_id}, and run it via \code{benchmark()}.
The resulting ROC curves are displayed in \Cref{fig:roc-curve} using \pkg{mlr3viz} \citep{ref-mlr3viz2025, ref-ggplot2}.
Again, we use only $10\%$ of the data to allow reproducing the code more easily.

\begin{CodeChunk}
\begin{CodeInput}
R> library("mlr3learners")
R> baseline <- as_learner(
+    po("select",
+      selector = selector_invert(selector_type("lazy_tensor"))) %>>%
+    po("imputehist") %>>%
+    lrn("classif.log_reg", predict_type = "prob"))
R> baseline$id <- "tabular"
R> glrn$id <- "multimodal"
R> task_subset <- task$clone(deep = TRUE)
R> ids <- partition(task_subset, ratio = 0.1)$train
R> task_subset$filter(ids)
R> design <- benchmark_grid(task_subset, list(glrn, baseline),
+    rsmp("holdout"))
R> bmr <- benchmark(design)
R> library("mlr3viz")
R> autoplot(bmr, type = "roc")
\end{CodeInput}
\end{CodeChunk}

Note that the ROC curves in \Cref{fig:roc-curve} are subject to high uncertainty because we are using little data and the problem is highly imbalanced.
This also explains why the ROC curve of the tabular baseline partly lies below the diagonal.

\section{Runtime evaluation}\label{sec:benchmarks}

The primary goal of this work is to provide a high-level interface to DL in \rlang{}, and the main computational tasks are handled by \torch{}.
The package primarily influencing the runtime performance is therefore \torch{} and not \pkg{mlr3torch}.
Nonetheless, training speed is an important consideration when deciding whether to adopt a package.
Furthermore, we want to ensure that \mlrttorch{} does not add a significant framework overhead when compared with a pure \torch{} implementation.
In this section, we present runtime benchmarks that compare a simple training loop using \mlrttorch{} with an equivalent implementation in pure \torch{} and \pytorch{}.
We include \pytorch{} in the evaluation scenario because, due to its popularity and heavy industry investment, it can reasonably be considered a best-case scenario for the runtime performance of a DL framework.
Despite \torch{} and \pytorch{} being built upon the same \libtorch{} \cpp{} library, there is still a considerable engineering effort in the \pytorch{} ``frontend'' to make it run as fast as it does.

\subsection{Setup}
One main reason for the success of DL is the use of GPUs for training neural networks, but small networks, for example, for tabular data, can also be trained on CPUs.
We therefore evaluate the runtime performance on both hardware backends.

We also vary several aspects of the training setup, where the main computational steps are
\begin{itemize}
    \item the forward pass of the neural network, i.e., computing predictions for a given feature vector, as well as the loss calculation,
    \item computing the gradients of the loss function with respect to the network's parameters, and
    \item performing the optimization step to update the network weights.
\end{itemize}

We train a simple MLP with ReLU activation and with different numbers of layers and latent dimensions.
Note that the number of floating-point operations (FLOPs) scales linearly with the number of layers, but quadratically with the latent dimension.
Moreover, increasing the number of layers causes the creation of more intermediate tensors that are managed by \rlang{} and have to eventually be cleaned up by the \rlang{} garbage collector.

We also use two optimizers: a simple SGD and the more computationally intensive (but more commonly used) AdamW.
In order to narrow the performance gap between \torch{} and \pytorch{}, we have contributed various improvements to \torch{}, most notably the addition of ``ignite'' optimizers in \torch{} version 0.14.0.
Instead of implementing the optimization step in \rlang{}, they wrap the \cpp{} implementation of the optimizers provided in \libtorch{}.

Each training loop runs for 20 epochs on mini-batches from a synthetically generated regression dataset with 2000 observations and 1000 features.
Note that we do not vary the number of features because it only influences the computation in the first layer.
We also train the network for four epochs as a warm-up in order to mitigate the impact of constant overheads on our measurements, ensuring that our reported runtimes more accurately reflect long-running training scenarios.
Every configuration is repeated 10 times.
On the GPU, we use batches of size 32 and latent dimensions of 1000, 3000, and 9000.
On the CPU, we use 16 threads, batches of size 32, and latent dimensions of 800, 1600, and 3200.
The results for a single CPU thread with the same batch size and latent dimensions of 100, 200, and 400 are shown in \Cref{app:single-thread}.
Further computational details can be found in~\Cref{app:comp-details}.

\subsection{Results}

\begin{figure}[t!]
    \centering
    \includegraphics[width=\textwidth]{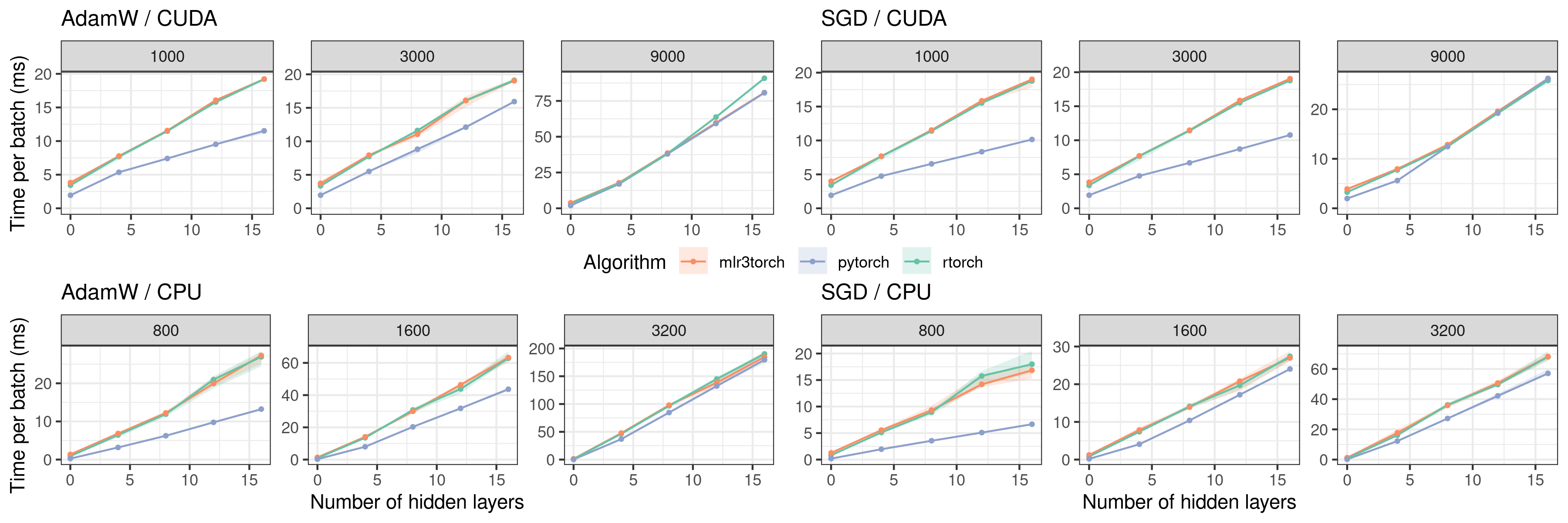}
    \caption{Runtime results for AdamW (left) and SGD (right) for GPU (top) and CPU with 16 threads (bottom). The facets correspond to different latent dimensions. The number of hidden layers is 0, 4, 8, 12, 16.
    The lines represent the median and the shaded area is defined by the 10\% and 90\% quantiles of the 10 repetitions (lower is better).
}
    \label{fig:optimizer-benchmark}
\end{figure}

\begin{figure}[t!]
\centering
    \includegraphics[width=\textwidth]{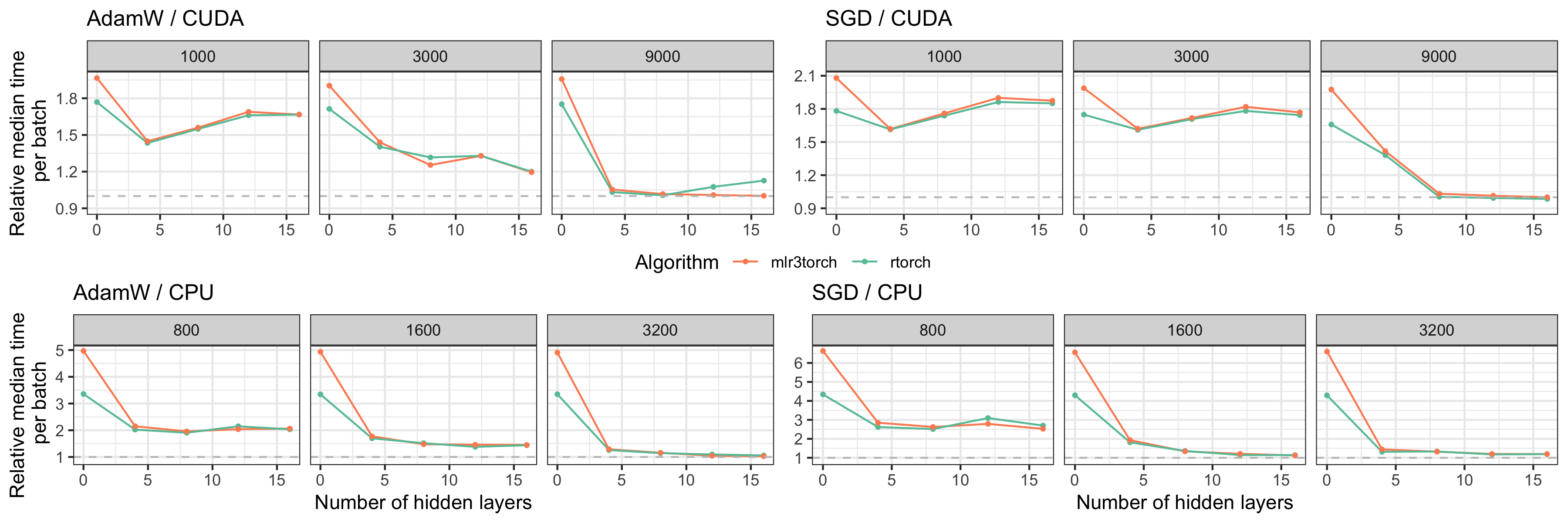}
    \caption{Median runtime of \mlrttorch{} and \torch{} relative to \pytorch{} (lower is better) for GPU (top) and CPU with 16 threads (bottom). The dashed line marks the runtime of \pytorch{}.}
    \label{fig:optimizer-benchmark-relative}
\end{figure}

We only show results for the ignite optimizers, as they are considerably faster than their corresponding \rlang{} implementations and are therefore used in \mlrttorch{}.
The difference between the two versions is especially pronounced for the more expensive AdamW optimizer, where, on the GPU, the relative overhead for a latent size of 1000 ranges from around 1.7 for 0 hidden layers to around 3.6 for 16 hidden layers.
For the same scenario using SGD, the overhead is only around 1.1.
Usually, however, more complex optimizers like AdamW are used to train neural networks.
\Cref{fig:optimizer-benchmark} displays the absolute time per batch and \Cref{fig:optimizer-benchmark-relative} shows relative results for the ignite optimizer version.

In all experiments, the results of \torch{} and \mlrttorch{} are relatively similar.
Only when the neural network is relatively small is the relative overhead more noticeable, which is expected, as \mlrttorch{} pays some runtime penalty for its generality.

When training on the GPU, the relative overhead of \torch{} with respect to \pytorch{} diminishes when increasing the latent dimension.
This is not always the case when increasing the number of hidden layers.
For example, for latent dimensions 1000 and 3000, the absolute overhead also increases with more layers, but when the network becomes very large, all three implementations are very similar.
Surprisingly, \mlrttorch{} is slightly faster than \torch{} for the AdamW optimizer when there are many layers and the latent dimension is 9000, but this is likely an artifact explained by some internal mechanism of \torch{} or \libtorch{}.

On the CPU with 16 threads, the relative overhead is largest for networks without hidden layers, where the fixed overhead per batch dominates and \mlrttorch{} is around 5 (AdamW) to 6.6 (SGD) times slower than \pytorch{}.
With hidden layers, the relative overhead of both \torch{} and \mlrttorch{} is considerably smaller for larger latent dimensions: for each number of hidden layers (4 to 16), it is around 2 (AdamW) and 2.5 to 3 (SGD) for a latent dimension of 800, and below 1.5 for a latent dimension of 3200.
Additionally, \Cref{fig:benchmark-single-thread} in \Cref{app:single-thread} shows results for a single CPU thread.

The benchmark also demonstrates the runtime advantage of the GPU over the CPU.
Note that the number of FLOPs increases quadratically with the latent dimension.
For small networks, the GPU is not faster than the CPU with 16 threads and is often slower, because the fixed overhead per batch dominates. For larger networks, the GPU is faster, even though its latent dimensions (1000, 3000, and 9000) are larger than those on the CPU (800, 1600, and 3200), and the advantage grows with the size of the network: accounting for the number of FLOPs, the GPU is roughly 20 times faster for the largest networks.

Overall, the runtime performance of \mlrttorch{} and \torch{} can be considered reasonable, especially when focusing on the arguably most realistic case of training a neural network on the GPU using the AdamW optimizer.

\section{Conclusion}\label{sec:conclusion}\nopagebreak %

In this work, we introduced \pkg{mlr3torch}, which equips \pkg{mlr3} with DL capabilities, so that neural networks can be resampled, tuned, and benchmarked like any other learner, and images and other tensors can be stored, preprocessed, and augmented lazily as columns of a task.
Its design offers different levels of control, from predefined architectures over \torch{} modules converted to learners to custom architectures in the graph language of \mlrtpipelines{}, and all of its components can be extended, e.g., through callbacks or custom task types.
The use cases showed how this supports a simple neural architecture search, the fine-tuning of pretrained image models, and multimodal learning, and the benchmarks demonstrated a reasonable runtime overhead of \mlrttorch{} compared to a pure \torch{} implementation.

Future development will build on the extension interfaces of \mlrttorch{} to provide learners for additional task types, such as survival analysis, alongside new architectures and further performance improvements.
Another direction is a closer integration with the Hugging Face Hub.
The \pkg{mlr3hf} package \citep{ref-mlr3hf}, which is currently in development, makes datasets from the Hugging Face Hub available as \mlrt{} tasks, and could in the future be extended to provide seamless access to pretrained models from the Hub within \mlrttorch{}.

\section*{Acknowledgments}

Sebastian Fischer is supported by the Deutsche Forschungsgemeinschaft (DFG, German Research Foundation) -- Project number 460135501 (NFDI project MaRDI).
We also thank Florian Pfisterer for fruitful discussions on the development of the package.

\bibliography{refs}

\begin{appendix}
\crefalias{section}{appendix} %

\newpage
\section{Reproducibility}\label{app:comp-details}

All relevant files for reproducing the results from the paper are provided in the \code{paper/} subdirectory of the software's GitHub repository: \url{https://github.com/mlr-org/mlr3torch}.
It contains a detailed \code{README.md} on how to reproduce the results.

For reproducibility purposes, we provide two Linux \pkg{Docker} images -- one for GPU (\proglang{CUDA}) and one for CPU -- that can be downloaded from Zenodo: \url{https://doi.org/10.5281/zenodo.23831328}.
They are primarily intended for reproducing the results from the runtime benchmark, as setting up \pytorch{} and \torch{} -- especially using \proglang{CUDA} -- can be tricky, but we also used the CPU \pkg{Docker} image for generating the results shown in the paper.
Note that \torch{} comes with its own BLAS implementation, so \rlang{} being installed with the default BLAS does not hurt performance.
The \code{paper/envs} directory in the Git repository contains the \code{Dockerfile}s, and the \code{paper} directory contains a \code{renv.lock} file that can be restored with \pkg{renv} \citep{ref-renv} to run the code without \pkg{Docker}.

To keep the runtime of the examples in this article short and easily reproducible with only a CPU, the paper uses small datasets and cheap configurations.
We also provide a slight variation of the code in the \code{paper} directory via the script \code{paper_code_expensive.R}.
It uses a GPU, the full datasets, and 100 instead of 20 tuning evaluations.
It also replaces the small \code{"mtcars"} example problem from the tuning section with the \code{"california_housing"} task \citep{ref-california-housing} with 20640 observations instead.

All experiments were run on Linux.
The CUDA benchmark was run on a single NVIDIA A100-SXM4 GPU with 80 GB of memory, which is part of an NVIDIA HGX A100 node with four GPUs.
This node had two Intel Xeon Gold 6336Y CPUs (2.40 GHz), i.e., 48 physical cores in total, and 1 TB RAM.
The CPU benchmark was run on a node of the CoolMUC-4 Linux cluster at the Leibniz Supercomputing Centre with two Intel Xeon Platinum 8480+ (Sapphire Rapids) CPUs, i.e., 112 physical cores in total, and 512 GB RAM.

We were able to exactly reproduce the results from the paper (excluding the runtime benchmark) throughout multiple runs on the same Linux machine using the CPU \pkg{Docker} image.
However, results that involve training a neural network can differ between machines, even when using the same \pkg{Docker} image.
This is the case not only between architectures (e.g., an M1 MacBook (ARM64) vs.\ x86-64), but also between x86-64 CPUs that support different instruction set extensions (e.g., AVX2 vs.\ AVX-512), because \libtorch{} selects different, numerically non-identical kernels depending on the available hardware.
For a discussion on reproducibility in \pytorch{}, which uses the same \libtorch{} library as \torch{}, see \url{https://docs.pytorch.org/docs/stable/notes/randomness.html\#reproducibility}.
The runtime benchmark is not exactly reproducible, but repetitions on the same hardware yielded rather similar results with some minor differences.
When run on different hardware, the differences in runtime measurements were non-negligible, but still compatible with the conclusions drawn in this paper.

\section{Single-threaded CPU results}\label{app:single-thread}

\begin{figure}[t!]
\centering
    \includegraphics[width=\textwidth]{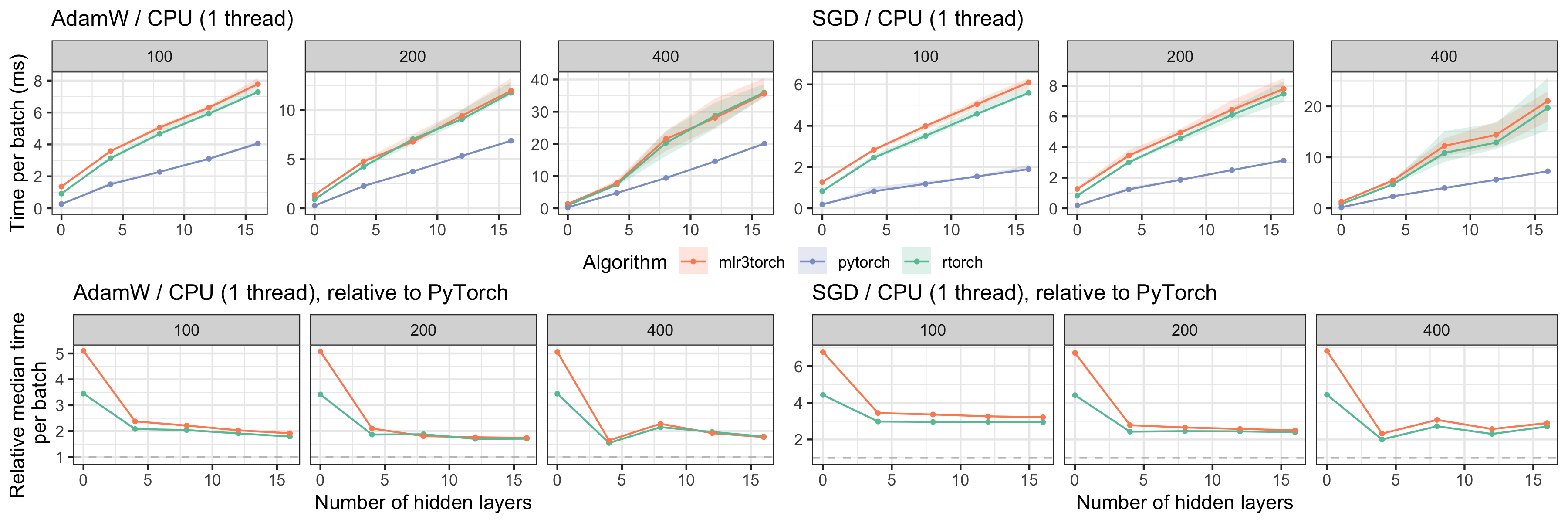}
    \caption{Runtime results on the CPU with a single thread for AdamW (left) and SGD (right): absolute time per batch (top) and median runtime relative to \pytorch{} (bottom; lower is better). The facets correspond to different latent dimensions, the shaded areas are defined by the 10\% and 90\% quantiles of the 10 repetitions, and the dashed line marks the runtime of \pytorch{}.}
    \label{fig:benchmark-single-thread}
\end{figure}

For the benchmark with a single CPU thread, we use smaller latent dimensions (100, 200, and 400) than for the benchmark with 16 threads (800, 1600, and 3200), as running the larger networks on a single thread would have taken too long.
The results are shown in \Cref{fig:benchmark-single-thread}.
As with 16 threads, \mlrttorch{} is around 5 to 6.8 times slower than \pytorch{} for networks without hidden layers, where the fixed overhead per batch dominates.
With hidden layers, it is around 1.6 to 3.5 times slower than \pytorch{}, and its runtime is within around 16\% of that of \torch{}.

\end{appendix}

\end{document}